\documentclass[10pt,journal,letterpaper,compsoc]{IEEEtran}
\usepackage{amsmath,amsfonts,amssymb,epsfig,verbatim,color,amsthm}
\usepackage{graphicx}
\usepackage{multirow}
\usepackage{rotating}
\usepackage{booktabs}
\usepackage{pdfpages}
\usepackage{subfig}
\usepackage{algorithm}
\usepackage{algpseudocode}
\usepackage{url}
\usepackage{bm}
\usepackage{array}
\usepackage{xcolor,colortbl}
\usepackage{textcomp}
\usepackage{multirow}

\newcommand{\mat}[1]{\mathtt #1}

\newcommand{\ind}[1]{\mathbb 1_{#1}}
\newcommand{\vct}[1]{\mathbf #1}

\newcommand{\argmin}{\operatornamewithlimits{\arg\,\min}}
\newcommand{\x}{{\bf x}}

\newcommand{\et}{\textit{et al. }}

\newtheorem{proposition}{Proposition}

\DeclareMathAlphabet{\mathpzc}{OT1}{pzc}{m}{it}

\begin{document}
	
	\title{Dominant Sets for ``Constrained'' Image Segmentation}
	
	\author{Eyasu Zemene,~\IEEEmembership{Member,~IEEE,}
		Leulseged Tesfaye Alemu,~\IEEEmembership{Member,~IEEE}
		and~Marcello~Pelillo,~\IEEEmembership{Fellow,~IEEE}
		\IEEEcompsocitemizethanks{\IEEEcompsocthanksitem The authors are with the Dipartimento di Scienze Ambientali, Informatica
			e Statistica, Universita Ca' Foscari Venezia, via Torino 155, 30172 Venezia Mestre, Italy. E-mail: \{eyasu.zemene, leuelseged, pelillo\}@unive.it}}

	\markboth{}%
	{Shell \MakeLowercase{\textit{et al.}}: Bare Demo of IEEEtran.cls for Computer Society Journals}
	
	\IEEEcompsoctitleabstractindextext{

		\begin{abstract}
			Image segmentation has come a long way since the early days of computer vision, and still remains a challenging task. Modern variations of the classical (purely bottom-up) approach, involve, e.g., some form of user assistance (interactive segmentation) or ask for the simultaneous segmentation of two or more images (co-segmentation). At an abstract level, all these variants can be thought of as ``constrained'' versions of the original formulation, whereby the segmentation process is guided by some external source of information. In this paper, we propose a new approach to tackle this kind of problems in a unified way. Our work is based on some properties of a family of quadratic optimization problems related to {\em dominant sets}, a well-known graph-theoretic notion of a cluster which generalizes the concept of a maximal clique to edge-weighted graphs. In particular, we show that by properly controlling a regularization parameter which determines the structure and the scale of the underlying problem, we are in a position to extract groups of dominant-set clusters that are constrained to contain predefined elements. 
			In particular, we shall focus on interactive segmentation and co-segmentation (in both the unsupervised and the interactive versions). The proposed algorithm can deal naturally with several type of constraints and input modality, including scribbles, sloppy contours, and bounding boxes, and is able to robustly handle noisy annotations on the part of the user. Experiments on standard benchmark datasets show the effectiveness of our approach as compared to state-of-the-art algorithms on a variety of natural images under several input conditions and constraints.
			
		\end{abstract}
		
		}
	
	\maketitle
	
	\IEEEdisplaynotcompsoctitleabstractindextext \IEEEpeerreviewmaketitle

	\section{Introduction}
	
	\IEEEPARstart{I}{mage} segmentation is arguably one of the oldest and best-studied problems in computer vision, being a fundamental step in a 
	variety of real-world applications, and yet remains a challenging task \cite{RichardSzeliski11} \cite{ForsPonse11}.
	Besides the standard, purely bottom-up formulation, which involves partitioning an input image into coherent regions, in the past few years several variants have been proposed which are attracting increasing attention within the community.
	Most of them usually take the form of a ``constrained'' version of the original problem, whereby the segmentation process is guided by some external source of information.
	
	For example, user-assisted (or ``interactive'') segmentation has become quite popular nowadays, especially because of its potential applications in problems such as image and video editing, medical image analysis, etc. \cite{GrabCutRotherKB04,iccvLempitsky09,MilCutCVPR14,BaiSapIJCV2009,LiSunTanShuACM2004,BoyJolICCV2001,MorBarIP1998}.
	Given an input image and some information provided by a user, usually in the form of a scribble or of a bounding box,
	the goal is to provide as output a foreground object in such a way as to best reflect the user's intent.
	By exploiting high-level, semantic knowledge on the part of the user, which is typically difficult to formalize, we are therefore able to effectively solve segmentation problems which would be otherwise too complex to be tackled using fully automatic segmentation algorithms.
	
	Existing algorithms fall into two broad categories, depending on whether the user annotation is given in terms of a scribble or of a bounding box, and supporters of the two approaches have both good reasons to prefer one
	modality against the other.
	For example, Wu et al. \cite{MilCutCVPR14} claim that bounding boxes are the most natural and economical form
	in terms of the amount of user interaction, and develop a multiple instance learning algorithm that extracts an arbitrary object located inside a tight bounding box at unknown location.
	Yu et al. \cite{LOOSECUTcorr15} also support the bounding-box approach, though their algorithm is different from others in that it does not need bounding boxes tightly enclosing the object of interest, whose production of course increases the annotation burden. They provide an algorithm, based on a Markov Random Field (MRF) energy function, that can handle input bounding box that only loosely covers the foreground object. 
	Xian et al. \cite{XiaZhaCheXuDinCoRR2015} propose a method which avoids the limitations of existing bounding box methods - region of interest (ROI) based methods, though they need much less user interaction, their performance is sensitive to initial ROI. 
	
	On the other hand, several researchers, arguing that boundary-based interactive segmentation such as intelligent scissors \cite{MorBarIP1998} requires the user to trace the whole boundary of the object, which is usually a time-consuming and tedious process, support scribble-based segmentation. Bai et al. \cite{BaiWuCVPR2014}, for example, propose a model based on ratio energy function which can be optimized using an iterated graph cut algorithm, which tolerates errors in the user input.
	In general, the input modality in an interactive segmentation algorithm affects both its accuracy and its ease of use. Existing methods work typically on a single modality and they focus on how to use that input most effectively. However, as noted recently by Jain and Grauman \cite{JainGraICCV2013}, sticking to one annotation form leads to a suboptimal tradeoff between human and machine effort, and they tried to estimate how much user input is required to sufficiently segment a novel input.

	
	Another example of a  ``constrained'' segmentation problem is co-segmentation. Given a set of images, the goal here is to jointly segment same or similar foreground objects. The problem was first introduced by Rother \et \cite{CarThoAndVlaCVPR2006} who used histogram matching to simultaneously segment the foreground object out from a given pair of images. Recently, several techniques have been proposed which try to co-segment groups containing more than two images, even in the presence of similar backgrounds. Joulin \et \cite{ArmFraJeaCVPR2010}, for example, proposed a discriminative clustering framework, combining normalized cut and kernel methods and the framework has recently been extended in an attempt to handle multiple classes and a significantly larger number of images \cite{ArmFraJeaCVPR2012}.

	The co-segmentation problem has also been addressed using user interaction \cite{DhrAdaDevJieTsuCVPR2010, XinJiaLinMinTIP2015}. Here, a user adds guidance, usually in the form of scribbles, on foreground objects of some of the input images. Batra \et \cite{DhrAdaDevJieTsuCVPR2010} proposed an extension of the (single-image) interactive segmentation algorithm of Boykov and Jolly \cite{BoyJolICCV2001}. They also proposed an algorithm that enables users to quickly guide the output of the co-segmentation algorithm towards the desired output via scribbles. Given scribbles, both on the background and the foreground, on some of the images, they cast the labeling problem as energy minimization defined over graphs constructed over each image in a group. Dong \et \cite{XinJiaLinMinTIP2015} proposed a method using global and local energy optimization. Given background and foreground scribbles, they built a foreground and a background Gaussian mixture model (GMM) which are used as global guide information from users. By considering the local neighborhood consistency, they built the local energy as the local smooth term which is automatically learned using spline regression. The minimization problem of the energy function is then converted into constrained quadratic programming (QP) problem, where an iterative optimization strategy is designed for the computational efficiency.

	
	In this paper (which is an extended version of \cite{ZemPelECCV16}), we propose a unified approach to address this kind of problems which can deal naturally with various type of input modality, or constraints, and is able to robustly handle noisy annotations on the part of the external source. In particular, we shall focus on interactive segmentation and co-segmentation (in both the unsupervised and the interactive versions).
	Our approach is based on some properties of a parameterized family of quadratic optimization problems related to dominant-set clusters, a well-known generalization of the notion of maximal cliques to edge-weighted graph which have proven to be extremely effective in a variety of computer vision problems, including (automatic) image and video segmentation \cite{PavPelCVPR2003,PavPel07} (see \cite{RotPel17} for a recent review).
	In particular, we show that by properly controlling a regularization parameter which determines the structure
	and the scale of the underlying problem, we are in a position to extract groups of dominant-set clusters which
	are constrained to contain user-selected elements. We provide bounds that allow us to control this process,
	which are based on the spectral properties of certain submatrices of the original affinity matrix.
	
	The resulting algorithm has a number of interesting features which distinguishes it from existing approaches.
	Specifically: 1) it is able to deal in a flexible manner with {\em both} scribble-based and
	boundary-based input modalities (such as sloppy contours and bounding boxes); 2) in the case of noiseless scribble inputs, it asks the user to provide {\em only} foreground pixels; 3) it turns out to be {\em robust} in the presence of input noise, allowing the user to draw, e.g., imperfect scribbles (including background pixels) or loose bounding boxes.

Experimental results on standard benchmark datasets demonstrate the effectiveness of our approach as compared to state-of-the-art algorithms on a wide variety of natural images under several input conditions.
	\begin{figure*}[t]
		\centering
		\includegraphics[width=1\linewidth,trim=0cm 8cm 0cm 0cm,clip ]{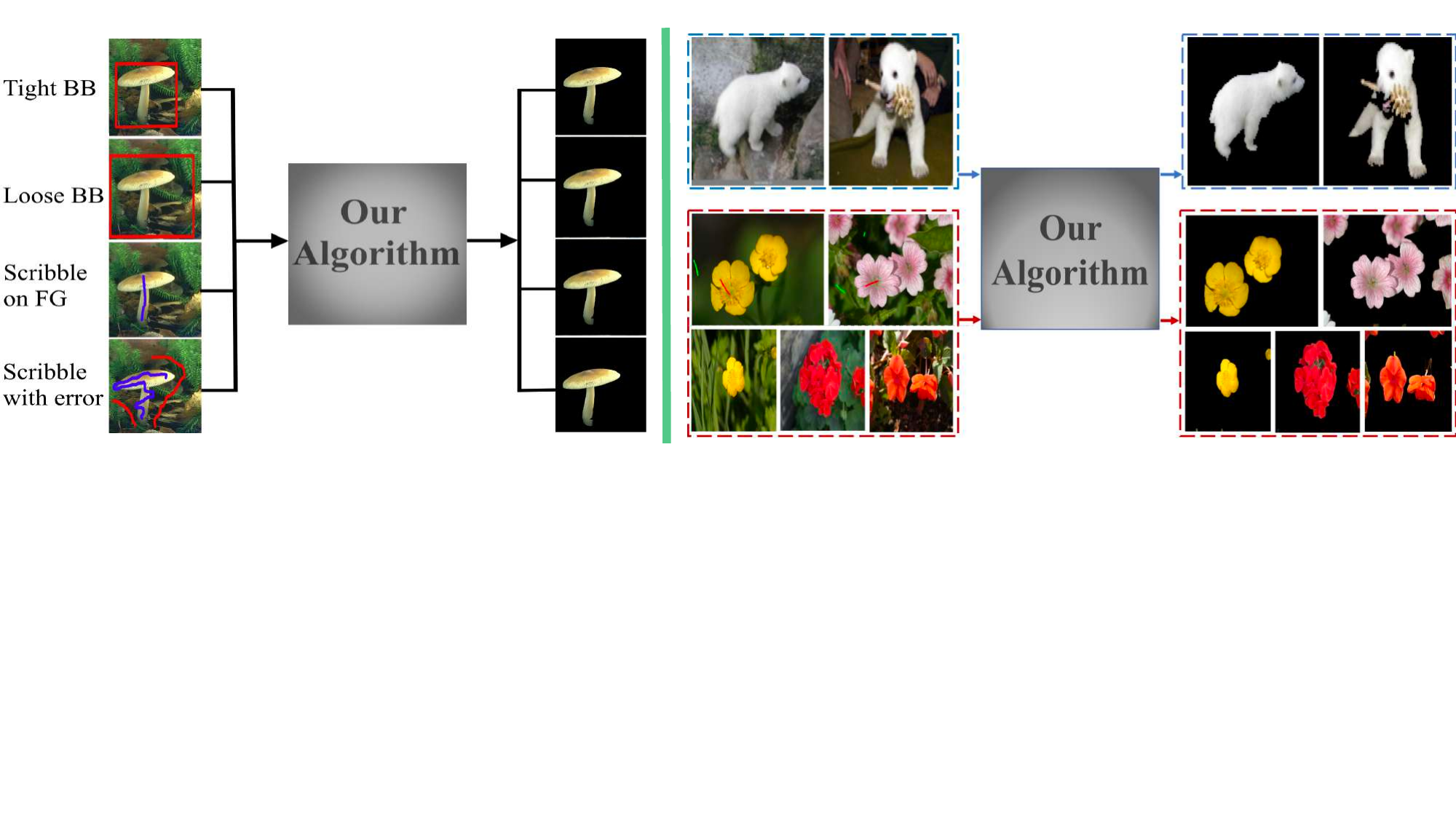}\\
		\caption{\small{ \bf Left:} An example of our interactive image segmentation method and its outputs, with different user annotation. Respectively from top to bottom, tight bounding box (Tight BB), loose bounding box (Loose BB), a scribble made (only) on the foreground object (Scribble on FG) and scribbles with errors. \textbf{Right:} Blue and Red dash-line boxes, show an example of our unsupervised and interactive co-segmentation methods, respectively. }
		\label{fig:InputModalities}
	\end{figure*}
Figure \ref{fig:InputModalities} shows some examples of how our system works in both interactive segmentation, in the presence of different input annotations, and co-segmentation settings.

\section{Dominant sets and quadratic optimization}
\label{sec:DS}
In the dominant set framework, the data to be clustered are represented as an undirected edge-weighted graph with no self-loops $G = (V, E,w)$, where $V = \{1, . . . , n\}$ is the vertex set, $E \subseteq V \times V$ is the edge set, and $w : E \rightarrow R_+^*$ is the (positive) weight function. Vertices in $G$ correspond to data points, edges represent neighborhood relationships, and edge-weights reflect similarity between pairs of linked vertices. As customary, we represent the graph $G$ with the corresponding weighted adjacency (or similarity) matrix, which is the $n \times n$ nonnegative, symmetric matrix $A = (a_{ij})$ defined as $a_{ij} = w(i, j)$, if $(i, j) \in E$, and $a_{ij} = 0$ otherwise. Since in $G$ there are no self-loops, note that all entries on the main diagonal of $A$ are zero.

For a non-empty subset $S \subseteq V$, $i \in S$, and $j \notin S$, define
\begin{equation}
\label{eq1}
\phi_S(i,j)=a_{ij}-\frac{1}{|S|} \sum_{k \in S} a_{ik}~.
\end{equation}
This quantity measures the (relative) similarity between nodes $j$ and $i$, with respect to the average similarity between node $i$ and its neighbors in $S$. Note that $\phi_S(i,j)$ can be either positive or negative. Next, to each vertex $i \in S$ we assign a weight defined (recursively) as follows:
\begin{equation}
w_S(i)=
\begin{cases}
1,&\text{if\quad $|S|=1$},\\
\sum_{j \in S \setminus \{i\}} \phi_{S \setminus \{i\}}(j,i)w_{S \setminus \{i\}}(j),&\text{otherwise}~.
\end{cases}
\end{equation}
Intuitively, $w_S(i)$ gives us a measure of the overall similarity between vertex $i$ and the vertices of $S\setminus \{i\}$
with respect to the overall similarity among the vertices in $S\setminus \{i\}$. Therefore, a positive $w_S(i)$ indicates that adding $i$ into its neighbors in $S$ will increase the internal coherence of the set, whereas in the presence of a negative value we expect the overall coherence to be decreased. Finally, the total weight of $S$ can be simply defined as
\begin{equation}
W(S)=\sum_{i \in S}w_S(i)~.
\end{equation}

A non-empty subset of vertices $S \subseteq V$ such that $W(T) > 0$ for any non-empty $T \subseteq S$, is said to be a {\em dominant set} if:
\begin{enumerate}
\item $w_S(i)>0$, for all $i \in S$,
\item $w_{S \cup \{i\}}(i)<0$, for all $i \notin S$.
\end{enumerate}
It is evident from the definition that a dominant set satisfies the two basic properties of a cluster: internal coherence and external incoherence. Condition 1 indicates that a dominant set is internally coherent, while condition 2 implies that
this coherence will be destroyed by the addition of any vertex from outside. In other words, a dominant set is a maximally coherent data set.

Now, consider the following linearly-constrained quadratic optimization problem:
\begin{equation}
\label{eq2}
\begin{array}{ll}
   \text{maximize }  &  f(\x) = \x' A \x \\
   \text{subject to} &  \mathbf{x} \in \Delta
\end{array}
\end{equation}
where a prime denotes transposition and  
$$
\Delta=\left\{ \x \in R^n~:~ \sum_{i=1}^n x_i = 1, \text{ and } x_i \geq 0 \text{ for all } i=1 \ldots n \right\}
$$ 
is the standard simplex of $R^n$.
In \cite{PavPelCVPR2003,PavPel07} a connection is established between dominant sets and the local solutions of \eqref{eq2}. In particular, it is shown that if $S$ is a dominant set then its ``weighted characteristics vector,'' which is the vector of $\Delta$ defined as,
\begin{displaymath}
x_i=
\begin{cases} \frac{w_S(i)}{W(s)},&\text{if\quad $i \in S$},\\ 0,&\text{otherwise}
\end{cases}
\end{displaymath}
is a strict local solution of \eqref{eq2}. Conversely, under mild conditions, it turns out that if $\x$ is a (strict) local solution of program \eqref{eq2} then its ``support''
$$
\sigma(\x) = \{i \in V~:~x_i > 0\}
$$ 
is a dominant set.
By virtue of this result, we can find a dominant set by first localizing a solution of program \eqref{eq2} with an appropriate continuous optimization technique, and then picking up the support set of the solution found. In this sense, we indirectly perform combinatorial optimization via continuous optimization. A generalization of these ideas to hypergraphs has recently been developed in \cite{RotPelPAMI2013}.

\begin{figure*}
	\centering
	\includegraphics[width=1\linewidth,height=0.2\linewidth]{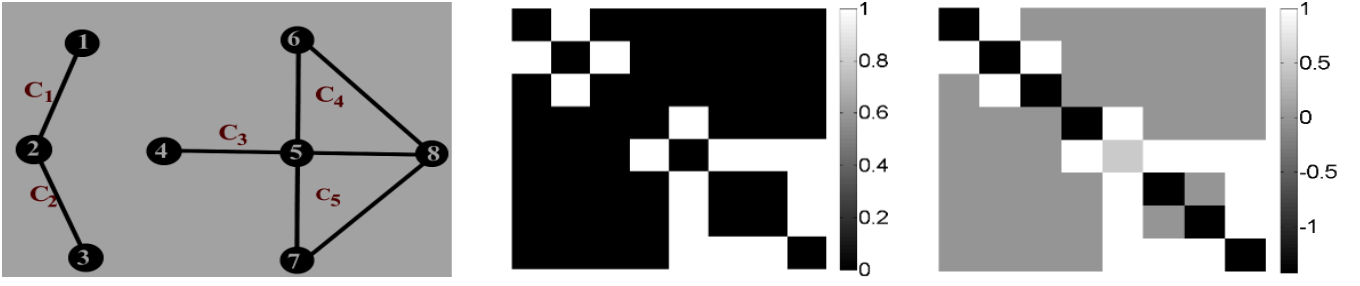}
	\caption{\small An example graph (left), corresponding affinity matrix (middle), and scaled affinity matrix 
		built considering vertex 5 as a user constraint (right). Notation $C_i$ refers to the $i^{th}$ maximal clique.}
	\label{fig:ExamplarGraphAndAffinity}
\end{figure*}

\section{Constrained dominant sets}

Let $G=(V,E,w)$ be an edge-weighted graph with $n$ vertices and let $A$ denote as usual its (weighted) adjacency matrix. Given a subset of vertices $S \subseteq V$ and a parameter $ \alpha > 0$, define
the following parameterized family of quadratic programs:
\begin{equation}
\label{eqn:parQP}
\begin{array}{ll}
   \text{maximize }  &  f_S^\alpha(\x) = \x' (A - \alpha \hat I_S) \x \\
   \text{subject to} &  \mathbf{x} \in \Delta
\end{array}
\end{equation}
where $\hat I_S$ is the $n \times n$ diagonal matrix whose diagonal elements are 
set to 1 in correspondence to the vertices contained in $V \setminus S$ and to zero otherwise, and the 0's
represent null square matrices of appropriate dimensions.
In other words, assuming for simplicity that $S$ contains, say, the first $k$ vertices of $V$, we have:

$$
\hat I_S = 
\begin{pmatrix} 
  ~~0~~ & ~~0~~ \\ 
  ~~0~~ & I_{n-k}  
\end{pmatrix}
$$
where $I_{n-k}$ denotes the $(n-k) \times (n-k) $ principal submatrix of the $n \times n$ identity matrix $I$ 
indexed by the elements of $V \setminus S$.
Accordingly, the function $f_S^\alpha$ can also be written as follows:
$$
f_S^\alpha(\x) = \x' A \x - \alpha \x'_S \x_S
$$
$\x_S$ being the $(n-k)$-dimensional vector obtained from
$\x$ by dropping all the components in $S$.
Basically, the function $f_S^\alpha$ is obtained from $f$ by inserting in the
affinity matrix $A$ the value of the parameter $\alpha$ in the main diagonal positions
corresponding to the elements of $V \setminus S$.

Notice that this differs markedly, and indeed
generalizes, the formulation proposed in \cite{PavPel03} for obtaining a hierarchical clustering 
in that here, only a subset of elements in the main diagonal
is allowed to take the $\alpha$ parameter, the other ones being set to zero.
We note in fact that the original (non-regularized) dominant-set formulation (\ref{eq2}) \cite{PavPel07} 
as well as its regularized counterpart described in \cite{PavPel03}
can be considered as degenerate version of ours, corresponding to the cases $S=V$ and $S=\emptyset$, respectively.
It is precisely this increased flexibility which allows us to use
this idea for finding groups of ``constrained'' dominant-set clusters.

We now derive the Karush-Kuhn-Tucker (KKT) conditions for program (\ref{eqn:parQP}),
namely the first-order necessary conditions for local optimality (see, e.g., \cite{Lue84}).
For a point $\x \in \Delta$ to be a KKT-point there should exist $n$
nonnegative real constants $\mu_1 , \ldots , \mu_n$ and an additional real number $\lambda$
such that 
\begin{displaymath}
[(A - \alpha \hat I_S) \x]_i - \lambda + \mu_i = 0
\end{displaymath}
for all $i=1 \ldots n$, and
\begin{displaymath}
\sum_{i=1}^n x_i \mu_i = 0~.
\end{displaymath}
Since both the $x_i$'s and the $\mu_i$'s are nonnegative, the
latter condition is equivalent to saying that $i \in \sigma(\x)$ implies
$\mu_i= 0$, from which we obtain:
\begin{displaymath}
[(A - \alpha \hat I_S) \x]_i ~
\begin{cases} 
~ = ~   \lambda, ~ \mbox{ if } i \in \sigma(\x) \\ 
~ \le ~ \lambda, ~ \mbox{ if } i \notin \sigma(\x)  
\end{cases}
\end{displaymath}
for some constant $\lambda$.
Noting that $\lambda = \x'A\x - \alpha \x'_S \x_S$ and recalling the definition of $\hat I_S$,
the KKT conditions can be explicitly rewritten as:
\begin{equation}
\label{eqn:KKT}
\left\{
\begin{array}{llll}
(A\x)_i - \alpha x_i  & =   & \x'A\x - \alpha \x'_S \x_S, & \mbox{ if } i \in \sigma(\x) \mbox{ and } i \notin S \\
(A\x)_i               & =   & \x'A\x - \alpha \x'_S \x_S, & \mbox{ if } i \in \sigma(\x) \mbox{ and } i \in S \\
(A\x)_i               & \le & \x'A\x - \alpha \x'_S \x_S, & \mbox{ if } i \notin \sigma(\x)
\end{array}
\right.
\end{equation}

We are now in a position to discuss the main results which motivate the algorithm presented in this paper.
Note that, in the sequel, given a subset of vertices $S\subseteq V$, the face of $\Delta$ corresponding to $S$ is given by: $\Delta_{S}=\{x\in \Delta : \sigma (x)\subseteq S\}$.

\begin{proposition}
\label{prop:gamma}
Let $S \subseteq V$, with $S \neq \emptyset$.
Define
\begin{equation}
\label{eqn:defgamma}
\gamma_S = \max_{ \x \in \Delta_{V \setminus S}} 
\min_{i \in S} ~ \frac{\x' A \x - (A\x)_i}{\x' \x}
\end{equation}
and let $\alpha > \gamma_S$.
If $\x$ is a local maximizer of $f_S^\alpha$ in $\Delta$, then
$\sigma(\x) \cap S \neq \emptyset$.
\end{proposition}

\proof
Let $\x$ be a local maximizer of $f_S^\alpha$ in $\Delta$, and suppose
by contradiction that no element of $\sigma(\x)$ belongs to $S$ or, in other words,
that $\x \in \Delta_{V \setminus S}$.
By letting 
$$
i = \argmin_{j \in S} ~ \frac{\x' A \x - (A\x)_j}{\x' \x}
$$
and observing that $\sigma(\x) \subseteq V \setminus S$ implies $\x' \x = \x'_S \x_S$,
we have: 
$$
\alpha >  \gamma_S \ge 
\frac{\x' A \x - (A\x)_i}{\x' \x} =
\frac{\x' A \x - (A\x)_i}{\x'_S \x_S}~.
$$
Hence, $(A\x)_i > \x' A \x - \alpha \x'_S \x_S$ for $i \notin \sigma(\x)$, 
but this violates the KKT conditions (\ref{eqn:KKT}), thereby proving the proposition.
\endproof

The following proposition provides a useful and easy-to-compute upper bound for $\gamma_S$. 

\begin{proposition}
\label{prop:bound}
Let $S \subseteq V$, with $S \neq \emptyset$. Then, 
\begin{equation}
\label{eqn:bound}
\gamma_S \le \lambda_{\max}(A_{V \setminus S})
\end{equation}
where $\lambda_{\max}(A_{V \setminus S})$ is the largest eigenvalue of the principal submatrix of $A$
indexed by the elements of $V \setminus S$. 
\end{proposition}
\proof
Let $\x$ be a point in $\Delta_{V \setminus S}$ which attains the maximum $\gamma_S$
as defined in (\ref{eqn:defgamma}).
Using the Rayleigh-Ritz theorem \cite{HorJon85} and the fact that $\sigma(\x)\subseteq V \setminus S$, we obtain:
$$
\lambda_{\max}(A_{V \setminus S}) \ge \frac{\x'_S A_{V \setminus S} \x_S}{\x'_S \x_S}
= \frac{\x' A \x}{\x' \x}~.
$$
Now, define $\gamma_S(\x) = \max \{ (A \x)_i~:~i \in S \}$.
Since $A$ is nonnegative so is $\gamma_S(\x)$, and recalling the definition of $\gamma_S$ we get:
$$
\frac{\x' A \x}{\x' \x} \ge \frac{\x' A \x - \gamma_S(\x)}{\x' \x} = \gamma_S
$$
which concludes the proof.
\endproof

The two previous propositions provide us with a simple technique to determine dominant-set clusters containing user-selected vertices. Indeed, if $S$ is the set of vertices selected by the user, by setting 
\begin{equation}
\label{alphabound}
\alpha > \lambda_{\max}(A_{V \setminus S})
\end{equation}
we are guaranteed that all local solutions of (\ref{eqn:parQP}) will have a support 
that necessarily contains elements of $S$.
%
%
Note that this does not necessarily imply that the (support of the) solution found corresponds to a dominant-set cluster of the original affinity matrix $A$, as adding the parameter $-\alpha$ on a portion of the main diagonal intrinsically changes the scale of the underlying problem. However, we have obtained extensive empirical evidence which supports a conjecture which turns out to be very useful for our interactive image segmentation application.

To illustrate the idea, let us consider the case where edge-weights are binary, which basically
means that the input graph is unweighted. In this case, it is known that dominant sets correspond to maximal cliques \cite{PavPel07}. Let $G=(V,E)$ be our unweighted graph and let $S$ be a subset of its vertices.
For the sake of simplicity, we distinguish three different situations of increasing generality.

\noindent
{\bf Case 1.} The set $S$ is a singleton, say $S = \{u\}$. In this case, we know from 
Proposition \ref{prop:bound} that all solutions $\x$ of 
$f_\alpha^S$ over $\Delta$ will have a support which contains $u$, that is $u \in \sigma(\x)$.
Indeed, we conjecture that there will be a unique local (and hence global) solution here whose support
coincides with the {\em union} of all maximal cliques of $G$ which contain vertex $u$. 

\noindent
{\bf Case 2.} The set $S$ is a clique, not necessarily maximal. In this case, 
Proposition \ref{prop:bound} predicts that all solutions $\x$ of (\ref{eqn:parQP})
will contain at least one vertex from $S$.
Here, we claim that indeed the support of local solutions is the union of the maximal cliques that contain $S$.

\noindent
{\bf Case 3.} The set $S$ is not a clique, but it can be decomposed as a collection of (possibly overlapping)
maximal cliques $C_1, C_2, ..., C_k$ (maximal with respect to the subgraph induced by $S$).
In this case, we claim that if $\x$ is a local solution, then its support can be obtained by taking the union of
all maximal cliques of $G$ containing one of the cliques $C_i$ in $S$.

To make our discussion clearer, consider the graph shown in Fig. \ref{fig:ExamplarGraphAndAffinity}. 
In order to test whether our claims hold, we used as the set $S$ different combinations of vertices, and
enumerated all local solutions of (\ref{eqn:parQP}) by multi-start replicator dynamics (see Section \ref{sec:DynamicsToExtractCDS}).
Some results are shown below, where on the left-hand side we indicate the set $S$, while on
the right hand-side we show the supports provided as output by the different runs of the algorithm. 
\begin{table}[h]
\begin{tabular}
{p{2.2cm}  p{0.5cm}  p{6cm}}
1.  ~~$S = \{ 2 \}$       &   $\Rightarrow$     & $\sigma(\x) = \{ 1, 2, 3 \}$ \\
2.  ~~$S = \{ 5 \}$       &   $\Rightarrow$     & $\sigma(\x) = \{ 4, 5, 6, 7, 8 \}$ \\
3.  ~~$S = \{ 4, 5 \}$    &   $\Rightarrow$     & $\sigma(\x) = \{ 4, 5 \}$ \\
4.  ~~$S = \{ 5, 8 \}$    &   $\Rightarrow$     & $\sigma(\x) = \{ 5, 6, 7, 8 \}$ \\
5.  ~~$S = \{ 1, 4 \}$    &   $\Rightarrow$     & $\sigma(\x_1) = \{ 1, 2 \}$, ~~$\sigma(\x_2) = \{ 4, 5 \}$\\
6.  ~~$S = \{ 2, 5, 8 \}$ &   $\Rightarrow$     & $\sigma(\x_1) = \{ 1, 2, 3 \}$, ~~$\sigma(\x_2) = \{ 5, 6, 7, 8 \}$
\end{tabular}
\end{table}

The previous observations can be summarized in the following general statement which does comprise all three cases. 
Let $S = C_1 \cup C_2 \cup \ldots \cup C_k$ ($k \ge 1$) be a subset of vertices of $G$, 
consisting of a collection of cliques $C_i$ ($i=1 \ldots k$).
Suppose that condition (\ref{alphabound}) holds, and let 
$\x$ be a local solution of (\ref{eqn:parQP}). Then, $\sigma(\x)$ consists of the union of 
all maximal cliques containing some clique $C_i$ of $S$.

We conjecture that the previous claim carries over to edge-weighted graphs
where the notion of a maximal clique is replaced by that of a dominant set.
In the supplementary material, we report the results of an extensive experimentation
we have conducted over standard DIMACS benchgraphs which provide support to our claim.
This conjecture is going to play a key role in our applications of these ideas
to interactive image segmentation.

\section{Finding constrained dominant sets using game dynamics}
\label{sec:DynamicsToExtractCDS}

Evolutionary game theory offers a whole class of simple dynamical systems to solve quadratic constrained optimization problems
like ours. It envisages a scenario in which pairs of players are repeatedly drawn at random from a large population of individuals to play a symmetric two-player game. Game dynamics are designed in such a way as to drive strategies with lower payoff to extinction, following 
Darwin's principle of natural selection \cite{Wei95,HofSig98}.

Let $x_i(t)$ is the proportion of the population which plays strategy $i$ $\in J$ (the set of strategies) at time $t$. The state of the population at any given instant is then given by $\vct{x}(t)$ = ($x_1(t),..., x_n(t)$)$'$ where $'$ denotes transposition and $n$ refers the size of available pure strategies, that is $|J|$. 

Let $W= (w_{ij})$ be the $n \times n$ payoff matrix (biologically measured as Darwinian fitness or as profits in economic applications). The payoff for the $i^{th}$-strategist, assuming the opponent is playing the $j^{th}$ strategy, is given by $w_{ij}$, the corresponding $i^{th}$ row and the $j^{th}$ column of $W$. If the population is in state $\vct{x}$, the expected payoff earned by an the $i^{th}$-strategist is:

\[\mathcal{P}_i(\vct{x}) = \sum\limits_{j=1}^{n}w_{ij}x_j = (W\vct{x})_i\]
and the mean payoff over the whole population is 
\[\mathcal{P}(\vct{x}) = \sum\limits_{i=1}^{n}x_i \mathcal{P}_i(\vct{x}) = \vct{x}'W\vct{x}\]

The game, which is assumed to be played over and over, generation after generation, changes the state of the population over time until equilibrium is reached. A point $\vct{x}$ is said to be a stationary (or equilibrium) point of the dynamical
system if $\dot{x}$ = 0 where the dot implies derivative with respect to time. 

Different formalization of this selection process have been proposed in evolutionary game theory. One of the best-known class of game dynamics is given by the so-called {\em replicator dynamics}, which prescribes that the average rate of increase $\dot{x}_i$/$x_i$ equals the difference between the average fitness of strategy $i$ and the mean fitness over the entire population:

\begin{equation}
	\dot{x} = x_i\left( (W\vct{x})_i - \vct{x}'W\vct{x} \right)
	\label{ContinuousReplicator}
\end{equation}

A well-known discretization of the above dynamics is:

\begin{equation}
	x_i^{(t+1)} = x_i^{(t)} \frac{(W\x^{(t)})_i}{(\x^{(t)})'W(\x^{(t)})}
	\label{eqn:Replicator}
\end{equation}

Now, the celebrated Fundamental Theorem of Natural Selection \cite{HofSig98} states that, if $W=W'$, then the
average population payoff $\vct{x}'W\vct{x}$ is strictly increasing along any non-constant trajectory of both the continuous-time and discrete-time replicator dynamics. Thanks to this property, replicator dynamics naturally suggest themselves as a simple heuristics for finding (constrained) dominant sets \cite{PavPel07}. 

In our case, problem (\ref{eqn:parQP}), the payoff matrix $W$ is given by

$$
W = A - \alpha \hat I_S
$$
which yields:

\begin{equation}
	\label{eqn:Replicator}
x_i^{(t+1)} ~=~
\begin{cases} 
x_i^{(t)} \frac{(A\x^{(t)})_i}{(\x^{(t)})'(A-\alpha \hat I_S)(\x^{(t)})}, ~ \mbox{ if } i \in S  \\
\\
x_i^{(t)} \frac{(A\x^{(t)})_i  - \alpha x_i^{(t)}}{(\x^{(t)})'(A-\alpha \hat I_S)(\x^{(t)})}, ~ \mbox{ if } i \notin S 
\end{cases}
\end{equation}

Provided that the matrix $A - \alpha \hat I_S$ is scaled properly to
avoid negative values, it is readily seen that the simplex $\Delta$
is invariant under these dynamics, which means that every
trajectory starting in $\Delta$ will remain in $\Delta$ for all future times.

Although in the experiments reported in this paper we used the replicator dynamics described above, we mention
a faster alternative to solve linearly constrained quadratic optimization problems like ours, namely {\em Infection and Immunization Dynamics} (InImDyn) \cite{RotPelBomCVIU2011}.
Each step of InImDyn has a linear time/space complexity as opposed to the quadratic per-step complexity of replicator dynamics, and is 
therefore to be preferred in the presence of large payoff matrices.

\section{Application to interactive image segmentation}

In this section, we apply our model to the interactive image segmentation problem. As input modalities we consider scribbles as well as boundary-based approaches (in particular, bounding boxes) and, in both cases,
we show how the system is robust under input perturbations, namely imperfect scribbles or loose bounding boxes.

In this application the vertices of the underlying graph $G$ represent the pixels of the input image (or superpixels, as discussed below), and the edge-weights reflect the similarity between them.
As for the set $S$, its content depends on whether we are using scribbles or bounding boxes 
as the user annotation modality. 
In particular, in the case of scribbles, $S$ represents precisely those pixels that have been
manually selected by the user. In the case of boundary-based annotation 
instead, it is taken to contain only the pixels comprising the box boundary,
which are supposed to represent the background scene.
Accordingly, the union of the extracted dominant sets, say $\mathcal{L}$ dominant sets are extracted which contain the set $S$,  as described in the previous section and below, $\mathbf{UDS}=\mathcal{D}_1 \cup \mathcal{D}_2 ..... \cup \mathcal{D}_{\mathcal{L}}$, represents either the foreground object or the background scene depending on the input modality. For scribble-based approach the extracted set, $\mathbf{UDS}$, represent the segmentation result, while in the boundary-based approach we provide as output the complement of the extracted set, namely $\mathbf{V}\setminus \mathbf{UDS}$. 

\begin{figure}
	\centering
	\includegraphics[width=1\linewidth, height=0.5\linewidth]{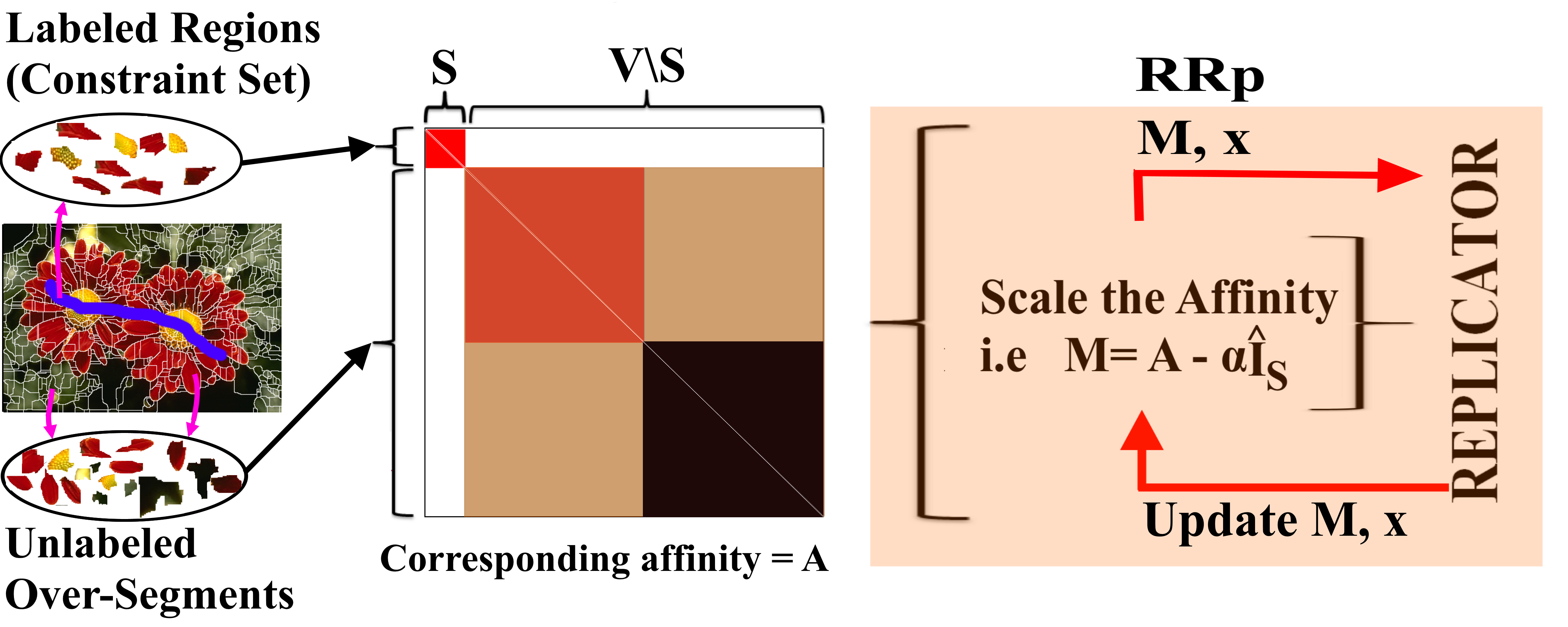}
	\caption{\small Overview of our interactive segmentation system. \textbf{Left:} Over-segmented image (output of the UCM-OWT algorithm \cite{MalikHierarchical}) with a user scribble (blue label). \textbf{Middle:} The corresponding affinity matrix, using each over-segments as a node, showing its two parts: $S$, the constraint set which contains the user labels, and $V\setminus S$, the part of the graph which takes the regularization parameter $\alpha$. \textbf{Right:} RRp, starts from the barycenter and extracts the first dominant set and update $\mathbf{x}$ and $\mathbf{M}$, for the next extraction till all the dominant sets which contain the user labeled regions are extracted.}
	\label{fig:Framework}
\end{figure}

Figure \ref{fig:Framework} shows the pipeline of our system. 
Many segmentation tasks reduce their complexity by using superpixels (a.k.a. over-segments) as a preprocessing step \cite{MilCutCVPR14,LOOSECUTcorr15,HoiEfrHebICCV2005} \cite{WanJiaHuaZhaQuaCVPR2008,XiaQuaICCV2009}. 
While \cite{MilCutCVPR14} used SLIC superpixels \cite{SLIC-superpixels-TPAMI12}, \cite{LOOSECUTcorr15} used a recent superpixel algorithm \cite{ZhoJuWanWACV2015} which considers not only the color/feature information but also boundary smoothness among the superpixels. 
In this work, we used the over-segments obtained from Ultrametric Contour Map (UCM) which is constructed from Oriented Watershed Transform (OWT) using globalized probability of boundary (gPb) signal as an input \cite{MalikHierarchical}.

We then construct a graph $G$ where the vertices represent over-segments and 
the similarity (edge-weight) between any two of them is obtained using a standard Gaussian kernel
$$\mat A^\sigma_{ij}=\ind{i\neq j}exp(\Vert\vct f_i-\vct f_j\Vert^2/{2\sigma^2})$$
where $\vct f_i$, is the feature vector of the $i^{th}$ over-segment, $\sigma$ is the free scale parameter, and $\ind{P}=1$ if $P$ is true, $0$ otherwise.

Given the affinity matrix $A$ and the set $S$ as described before, the system constructs
the regularized matrix $M=A-\alpha \hat I_S$, with $\alpha$ chosen as prescribed in (\ref{alphabound}).
Then, the replicator dynamics (\ref{eqn:Replicator}) are run (starting them as customary from the simplex barycenter) until they converge to some solution vector $\x$. We then take the support of $\x$, remove the
corresponding vertices from the graph and restart the replicator dynamics until all
the elements of $S$ are extracted.

\subsection{Experiments and results}

As mentioned above, the vertices of our graph represents over-segments
and edge weights (similarities) are built from the median of the color of all pixels in RGB, HSV, and L*a*b* color spaces, and Leung-Malik (LM) Filter Bank \cite{MalikLeungLMfilterIJCV01}. The number of dimensions of feature vectors for each over-segment is then 57 (three for each of the RGB, L*a*b*, and HSV color spaces, and 48 for LM Filter Bank).

In practice, the performance of graph-based algorithms that use Gaussian kernel, as we do, is sensitive to the selection of the scale parameter $\sigma$. In our experiments, we have reported three different results based on the way $\sigma$ is chosen: $1)$ CDS\_Best\_Sigma, in this case the best parameter $\sigma$  is selected on a per-image basis, which indeed can be thought of as the optimal result (or upper bound) of the framework. $2) $ CDS\_Single\_Sigma, the best parameter in this case is selected on a per-database basis tuning $\sigma$ in some fixed range, which in our case is between 0.05 and 0.2. $3)$ CDS\_Self\_Tuning, the $\sigma^2$ in the above equation is replaced, based on \cite{ManPieNIPS2004}, by $\sigma_i*\sigma_j$, where $\sigma_i = mean(KNN(f_i))$, the mean of the K\_Nearest\_Neighbor of the sample $f_i$, K is fixed in all the experiment as 7. 

\textbf{Datasets:} We conduct four different experiments on the well-known GrabCut dataset \cite{GrabCutRotherKB04} which has been used as a benchmark in many computer vision tasks \cite{LiMN13}\cite{iccvLempitsky09,TangECCV14,OneCutICCV13,MilCutCVPR14,LOOSECUTcorr15} \cite{PriMorCohCVPR2010,YanCaiZheLuoIP2010}. The dataset contains 50 images together with manually-labeled segmentation ground truth. The same bounding boxes as those in \cite{iccvLempitsky09} is used as a baseline bounding box. We also evaluated our scribbled-based approach using the well known Berkeley dataset which contains 100 images.

\textbf{Metrics:}
We evaluate the approach using different metrics: error rate, fraction of misclassified pixels within the bounding box, Jaccard index which is given by, following \cite{EvaluationMetricsMcGuinnessO10}, $J$ = $\frac{|GT \cap O|}{|GT \cup O|}$, where $GT$ is the ground truth and $O$ is the output. The third metric is the Dice Similarity Coefficient ($DSC$), which measures the overlap between two segmented object volume, and is computed as $DSC = \frac{2*|GT \cap O|}{|GT|+|O|}$.

\textbf{Annotations:} 
In interactive image segmentation, users provide annotations which guides the segmentation.
A user usually provides information in different forms such as scribbles and bounding boxes. The input modality affects both its accuracy and ease-of-use \cite{JainGraICCV2013}. However, existing methods fix themselves to one input modality and focus on how to use that input information effectively. This leads to a suboptimal tradeoff in user and machine effort. Jain et al. \cite{JainGraICCV2013} estimates how much user input is required to sufficiently segment a given image. In this work as we have proposed an interactive framework, figure \ref{fig:InputModalities}, which can take any type of input modalities we will use four different type of annotations: bounding box, loose bounding box, scribbles - only on the object of interest -, and scribbles with error as of \cite{BaiWuCVPR2014}.

\subsubsection{Scribble based segmentation} \label{Sec.Scribble}
Given labels on the foreground as constraint set, we built the graph and collect (iteratively) all unlabeled regions (nodes of the graph) by extracting dominant set(s) that contains the constraint set (user scribbles). We provided quantitative comparison against several recent state-of-the-art interactive image segmentation methods which uses scribbles as a form of human annotation: \cite{BoyJolICCV2001}, Lazy Snapping \cite{LiSunTanShuACM2004}, Geodesic Segmentation \cite{BaiSapIJCV2009}, Random Walker \cite{GraPAMI2006}, Transduction \cite{DucAudKerPonFloCVPR2008} , Geodesic Graph Cut \cite{PriMorCohCVPR2010}, Constrained Random Walker \cite{YanCaiZheLuoIP2010}.

We have also compared the performance of our algorithm againts Biased Normalized Cut (BNC) \cite{MajiNishVishMalCVPR11}, an extension of normalized cut, which incorporates a quadratic constraint (bias or prior guess) on the solution $\x$, where the final solution is a weighted combination of the eigenvectors of normalized Laplacian matrix. In our experiments we have used the optimal parameters according to \cite{MajiNishVishMalCVPR11} to obtain the most out of the algorithm.

Tables \ref{table:ScribblesResult},\ref{table:ScribblesResultBerkeley} and the plots in Figure \ref{fig:ExamplarResults} show the respective quantitative and the several qualitative segmentation results. Most of the results, reported on table \ref{table:ScribblesResult}, are reported by previous works \cite{LOOSECUTcorr15,MilCutCVPR14,iccvLempitsky09,PriMorCohCVPR2010,YanCaiZheLuoIP2010}.
We can see that the proposed CDS outperforms all the other approaches.

\begin{table}
	{
		\centering
		\begin{tabular}{l|r}
			
			Methods                                        & Error Rate \\ \hline
			BNC \cite{MajiNishVishMalCVPR11}                & 13.9 \\ \hline
			Graph Cut \cite{BoyJolICCV2001}                & 6.7 \\ \hline
			Lazy Snapping \cite{LiSunTanShuACM2004}        & 6.7 \\ \hline
			Geodesic Segmentation \cite{BaiSapIJCV2009}    & 6.8 \\ \hline
			Random Walker \cite{GraPAMI2006}               & 5.4 \\ \hline
			Transduction \cite{DucAudKerPonFloCVPR2008}    & 5.4 \\ \hline
			Geodesic Graph Cut \cite{PriMorCohCVPR2010}    & 4.8 \\ \hline
			Constrained Random Walker \cite{YanCaiZheLuoIP2010} & 4.1 \\ \hline
			CDS\_Self Tuning (Ours)               & \textbf{3.57} \\ \hline
			CDS\_Single Sigma (Ours)              & \textbf{3.80} \\ \hline
			CDS\_Best Sigma (Ours)                & 2.72 \\ \hline
		\end{tabular}
		\caption{\small Error rates of different scribble-based approaches on the Grab-Cut dataset.}
		\label{table:ScribblesResult}
	}
\end{table}

\begin{table}
	{
		\centering
		\begin{tabular}{l|r}
			Methods                          & Jaccard Index \\ \hline
			MILCut-Struct \cite{MilCutCVPR14}             & 84 \\ \hline
			MILCut-Graph \cite{MilCutCVPR14}              & 83 \\ \hline
			MILCut \cite{MilCutCVPR14}                    & 78 \\ \hline
			Graph Cut \cite{GrabCutRotherKB04}            & 77 \\ \hline
			Binary Partition Trees \cite{SalGarIP2000}    & 71 \\ \hline
			Interactive Graph Cut \cite{BoyJolICCV2001}   & 64 \\ \hline
			Seeded Region Growing \cite{AdaBisPAMI1994}   & 59 \\ \hline
			Simple Interactive O.E\cite{FriJanRojACM2005} & 63 \\ \hline
			CDS\_Self Tuning (Ours)                       & \textbf{93} \\ \hline
			CDS\_Single Sigma (Ours)                      & \textbf{93} \\ \hline
			CDS\_Best Sigma (Ours)                        & 95 \\ \hline
		\end{tabular}
		\caption{\small Jaccard Index of different approaches -- first 5 bounding-box-based -- on Berkeley dataset.}
		\label{table:ScribblesResultBerkeley}
	}
\end{table}

\textbf{Error-tolerant Scribble Based Segmentation.} This is a family of scribble-based approach, proposed by Bai et. al \cite{BaiWuCVPR2014}, which tolerates imperfect input scribbles thereby avoiding the assumption of accurate scribbles.  
We have done experiments using synthetic scribbles and compared the algorithm against recently proposed methods specifically designed to segment and extract the object of interest tolerating the user input errors \cite{BaiWuCVPR2014,LiuSunShuACM2009,OzaKemAydACM2012,SubParSolKauCGF2013}.

Our framework is adapted to this problem as follows. We give for our framework the foreground scribbles as constraint set and check those scribbled regions which include background scribbled regions as their members in the extracted dominant set. Collecting all those dominant sets which are free from background scribbled regions generates the object of interest.

\begin{figure*}
	\centering
	\includegraphics[width=1\linewidth,trim=0.15cm 10.0cm 7cm 0cm,clip ]{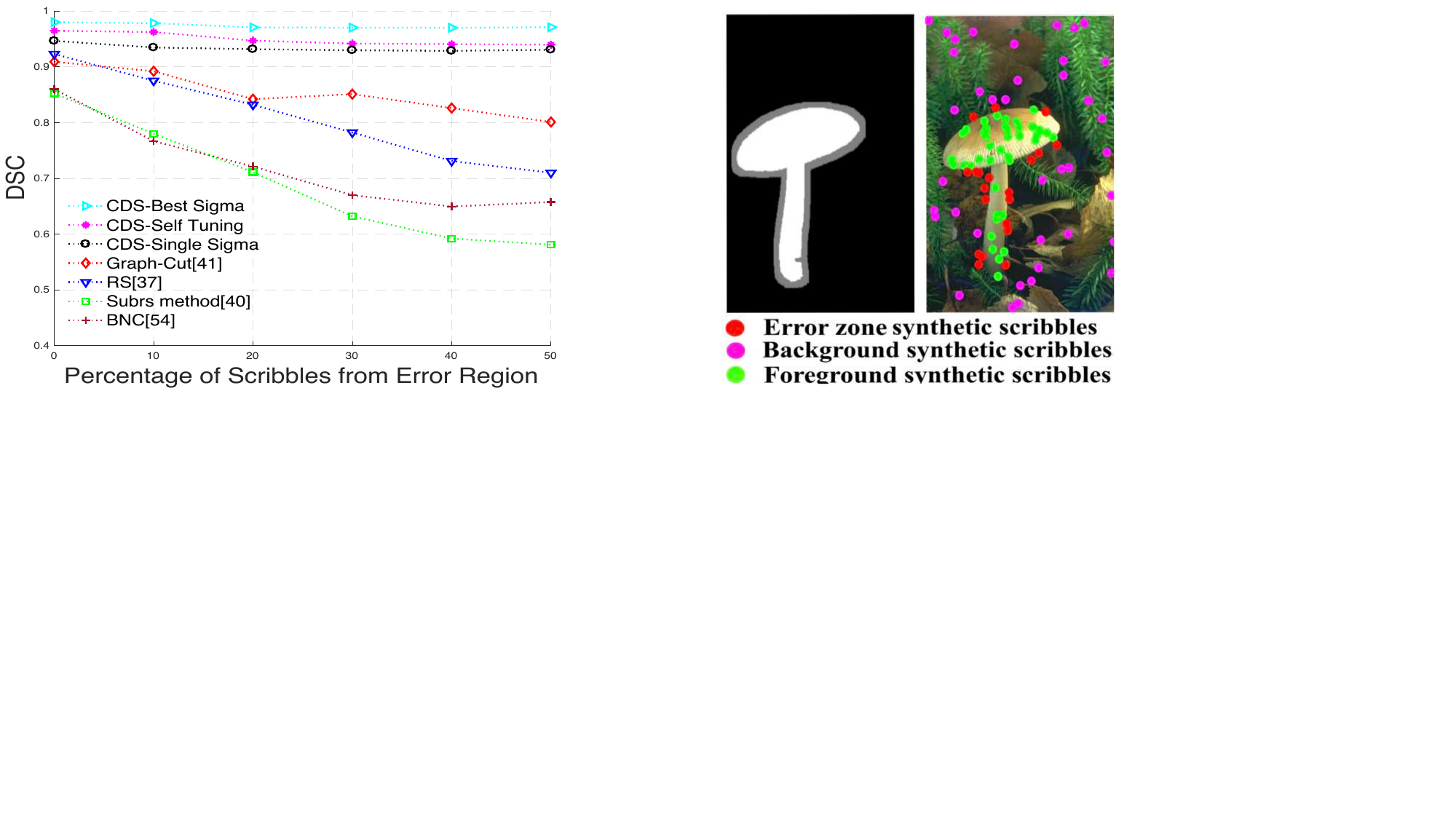}
	\caption{\small  \textbf{Left:} Performance of interactive segmentation algorithms, on Grab-Cut dataset, for different percentage of synthetic scribbles from the error region. \textbf{Right:} Synthetic scribbles and error region }
	\label{fig:SyntheticScribblesResult}
\end{figure*}

\textbf{Experiment using synthetic scribbles.}
Here, a procedure similar to the one used in \cite{SubParSolKauCGF2013} and \cite{BaiWuCVPR2014} has been followed.
First, 50 foreground pixels and 50 background pixels are
randomly selected based on ground truth (see Fig. \ref{fig:SyntheticScribblesResult}).
They are then assigned as foreground or background scribbles, respectively. Then
an error-zone for each image is defined as background pixels
that are less than a distance D from the foreground, in
which D is defined as 5 \%. We randomly select 0 to 50 pixels in the error zone and assign
them as foreground scribbles to simulate different degrees
of user input errors. We randomly select 0, 5, 10, 20, 30, 40, 50
erroneous sample pixels from error zone to simulate the error
percentage of 0\%, 10\%, 20\%, 40\%, 60\%, 80\%, 100\%
in the user input. It can be observed from figure \ref{fig:SyntheticScribblesResult} that our approach is not affected by the increase in the percentage of scribbles from error region.

\subsubsection{Segmentation using bounding boxes} \label{Sec.boundingBox}
The goal here is to segment the object of interest out from the background based on a given bounding box. The corresponding over-segments which contain the box label are taken as constraint set which guides the segmentation. The union of the extracted set is then considered as background while the union of other over-segments represent the object of interest.

\begin{figure*}[t]
	\centering
	\includegraphics[width=1\linewidth,trim=0cm 5.0cm 0cm 0cm,clip]{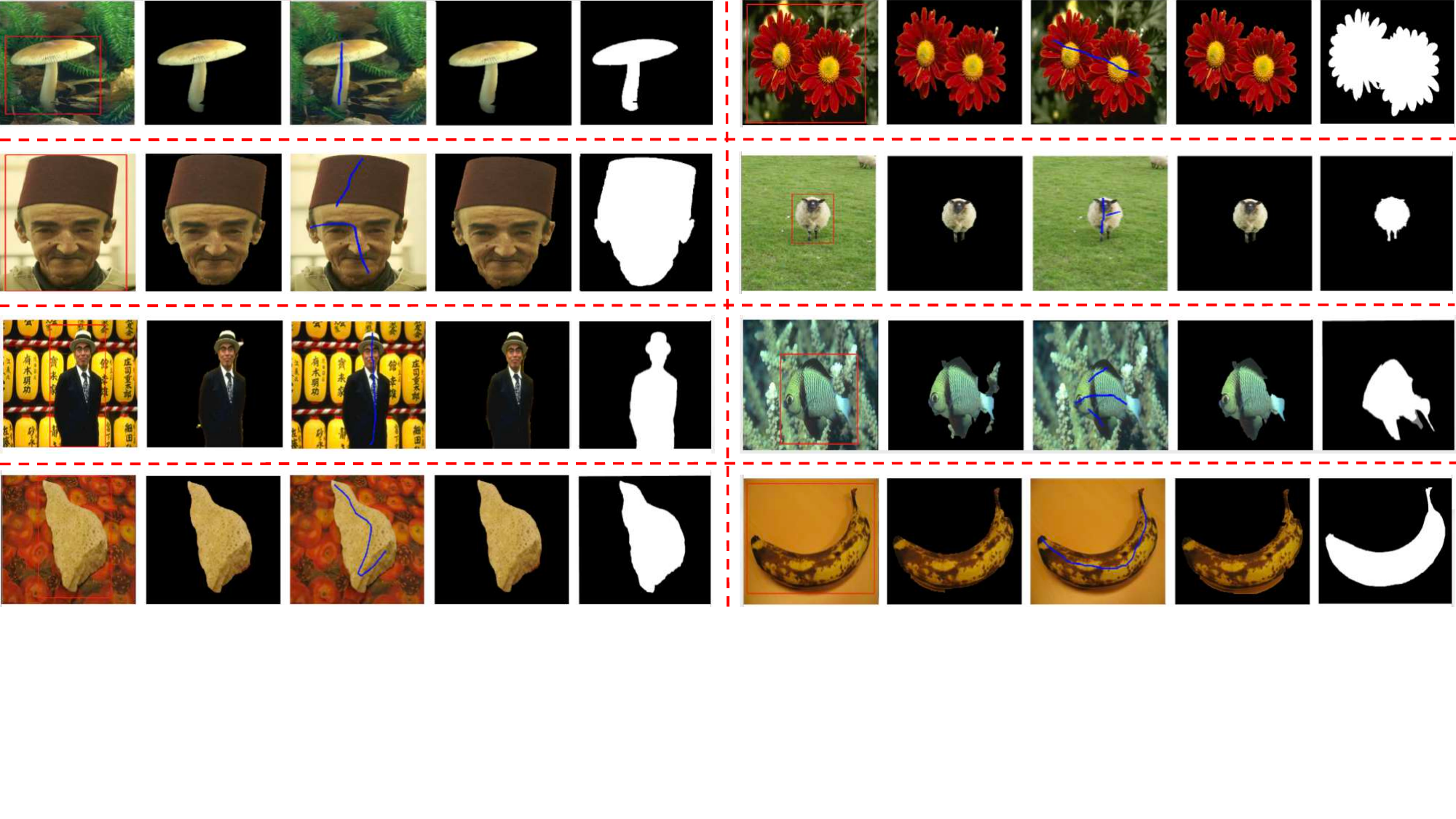}
	\caption{\small Examplar results of the interactive segmentation algorithm tested on Grab-Cut dataset. (In each block of the red dashed line) \textbf{Left:} Original image with bounding boxes of \cite{iccvLempitsky09}. \textbf{Middle left:} Result of the bounding box approach. \textbf{Middle:} Original image and scribbles (observe that the scribles are only on the object of interest). \textbf{Middle right:} Results of the scribbled approach. \textbf{Right:} The ground truth.}
	\label{fig:ExamplarResults}
\end{figure*}

We provide quantitative comparison against several recent state-of-the-art interactive image segmentation methods which uses bounding box: LooseCut \cite{LOOSECUTcorr15}, GrabCut \cite{GrabCutRotherKB04}, OneCut \cite{OneCutICCV13}, MILCut \cite{MilCutCVPR14}, pPBC and \cite{TangECCV14}.
Table \ref{table:LoosCutApproach} and the pictures in Figure \ref{fig:ExamplarResults} show the respective error rates and the several qualitative segmentation results. Most of the results, reported on table \ref{table:LoosCutApproach}, are reported by previous works \cite{LOOSECUTcorr15,MilCutCVPR14,iccvLempitsky09,PriMorCohCVPR2010,YanCaiZheLuoIP2010}.

\textbf{Segmentation Using Loose Bounding Box.} This is a variant of the bounding box approach, proposed by Yu et.al \cite{LOOSECUTcorr15}, which avoids the dependency of algorithms on the tightness of the box enclosing the object of interest. The approach not only avoids the annotation burden but also allows the algorithm to use automatically detected bounding boxes which might not tightly encloses the foreground object. It has been shown, in \cite{LOOSECUTcorr15}, that the well-known GrabCut algorithm \cite{GrabCutRotherKB04} fails when the looseness of the box is increased. Our framework, like \cite{LOOSECUTcorr15}, is able to extract the object of interest in both tight and loose boxes. Our algorithm is tested against a series of bounding boxes with increased looseness. The bounding boxes of \cite{iccvLempitsky09} are used as boxes with 0\% looseness. A looseness $L$ (in percentage) means an increase in the area of the box against the baseline one. The looseness is increased, unless it reaches the image perimeter where the box is cropped, by dilating the box by a number of pixels, based on the percentage of the looseness, along the 4 directions: left, right, up, and down.

For the sake of comparison, we conduct the same experiments as in \cite{LOOSECUTcorr15}: 41 images out of the 50 GrabCut dataset \cite{GrabCutRotherKB04} are selected as the rest 9 images contain multiple objects while the ground truth is only annotated on a single object. As other objects, which are not marked as an object of interest in the ground truth,  may be covered when the looseness of the box increases, images of multiple objects are not applicable
for testing the loosely bounded boxes \cite{LOOSECUTcorr15}. Table \ref{table:LoosCutApproach} summarizes the results of different approaches using bounding box at different level of looseness. As can be observed from the table, our approach performs well compared to the others when the level of looseness gets increased. When the looseness $L=0$, \cite{MilCutCVPR14} outperforms all, but it is clear, from their definition of tight bounding box, that it is highly dependent on the tightness of the bounding box. It even shrinks the initially given bounding box by 5\% to ensure its tightness before the slices of the positive bag are collected. For looseness of $L=120$ we have similar result with LooseCut \cite{LOOSECUTcorr15} which is specifically designed for this purpose. For other values of $L$ our algorithm outperforms all the approaches.

\begin{table}[t]
	\centering
	\begin{tabular}
		{p{3.2cm}    | p{0.8cm} | p{0.8cm}| p{0.8cm}| p{0.8cm}}
		\hline\hline
		Methods      & $ L = 0\% $ & $ L = 120\% $ & $ L = 240\% $ & $ L = 600\% $  \\ \hline
		GrabCut \cite{GrabCutRotherKB04}      & 7.4 &  10.1 &  12.6     &  13.7   \\ \hline
		OneCut \cite{OneCutICCV13}            & 6.6 & 8.7  &  9.9      &  13.7  \\ \hline
		pPBC \cite{TangECCV14}                & 7.5  &  9.1  &  9.4      &  12.3  \\ \hline
		MilCut \cite{MilCutCVPR14}            & \textbf{3.6}&  -    &  -        &  -       \\ \hline
		LooseCut \cite{LOOSECUTcorr15}        & 7.9 & \textbf{5.8}  &  6.9      &  6.8     \\ \hline
		CDS\_Self Tuning (Ours)               & 7.54 &  6.78 &  \textbf{6.35}      &  7.17 \\ \hline
		CDS\_Single Sigma (Ours)              & 7.48 &  5.9  &  \textbf{6.32}      & \textbf{6.29} \\ \hline
		CDS\_Best Sigma (Ours)                & 6.0  &  4.4  &  4.2     &  4.9 \\ \hline
	\end{tabular}
	\caption{\small Error rates of different bounding-box approaches with different level of looseness as an input, on the Grab-Cut dataset. $ L = 0\% $ implies a baseline bounding box as those in \cite{iccvLempitsky09}}
	\label{table:LoosCutApproach}
\end{table}

\textbf{Complexity:} 
In practice, over-segmenting and extracting features may be treated as a pre-processing step which can be done
before the segmentation process. Given the affinity matrix, we used replicator dynamics (\ref{eqn:Replicator}) to exctract
constrained dominant sets. Its computational complexity per step is $O(N^2)$, with $N$ being the total number of nodes of the graph.
Given that our graphs are of moderate size (usually less than 200 nodes) the algorithm is fast and converges 
in fractions of a second, with a code written in Matlab and run on a core i5 6 GB of memory. As for the pre-processing step, the original \textit{gPb-owt-ucm} segmentation algorithm was very slow to be used as a practical tools. Catanzaro et al. \cite{CatSuSunLeeMurKeuICCV2009} proposed a faster alternative, which reduce the runtime from 4 minutes to 1.8 seconds, reducing the  computational complexity and using parallelization which allow \textit{gPb} contour detector and \textit{gPb-owt-ucm} segmentation algorithm practical tools. For the purpose of our experiment we have used the Matlab implementation which takes around four minutes to converge, but in practice it is possible to give for our framework as an input, the GPU implementation \cite{CatSuSunLeeMurKeuICCV2009} which allows the convergence of the whole framework in around 4 seconds.

\section{Application to co-segmentation}

\begin{figure}
	\centering
	\includegraphics[width=1\linewidth ,trim=0cm 3cm 0cm 2cm,clip]{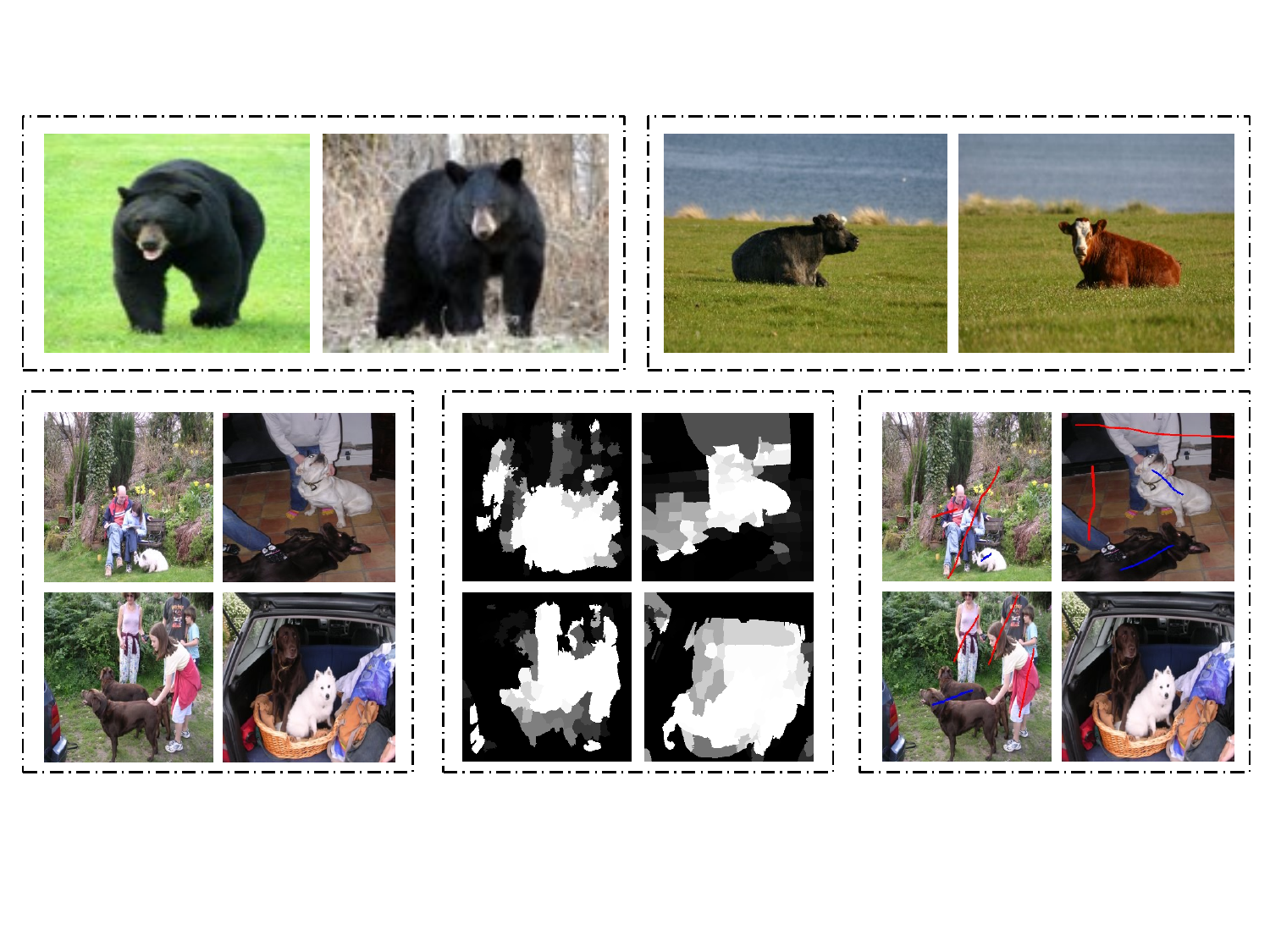}
	\caption{The challenges of co-segmentation. Examplar image pairs: \textbf{(top left)} similar foreground objects with significant variation in background, \textbf{(top right)} foreground objects with similar background. The \textbf{bottom} part shows why user interaction is important for some cases. The \textbf{bottom left} is the image, \textbf{bottom middle} shows the objectness score, and the \textbf{bottom right} shows the user label. }
	
	\label{fig:ImgPairExemplar}
\end{figure}

In this section, we describe the application of constrained dominant sets (CDS) to co-segmentation, both unsupervised and interactive. 
Among the difficulties that make this problem a challenging one, we mention the similarity among the different backgrounds and the similarity of object and background \cite{SarCarVlaCVPR2011} (see, e.g., the top row of Figure \ref{fig:ImgPairExemplar}). 
A measure of ``objectness'' has proven to be effective in dealing with such problems and improving the co-segmentation results \cite{SarCarVlaCVPR2011}\cite{AvikChaVelECCV16}. However, this measure alone is not enough, especially when one aims to solve the problem using global pixel relations. One can see from Figure \ref{fig:ImgPairExemplar} (bottom) that the color of the cloth of the person, which of course is one of the objects, is similar to the color of the dog which makes systems that are based on objectness measure fail. Moreover the object may not also be the one which we want to co-segment.

\begin{figure*}[t]
	\centering
	\includegraphics[width=1\linewidth ,trim=0cm 9.7cm 1cm 0.5cm,clip]{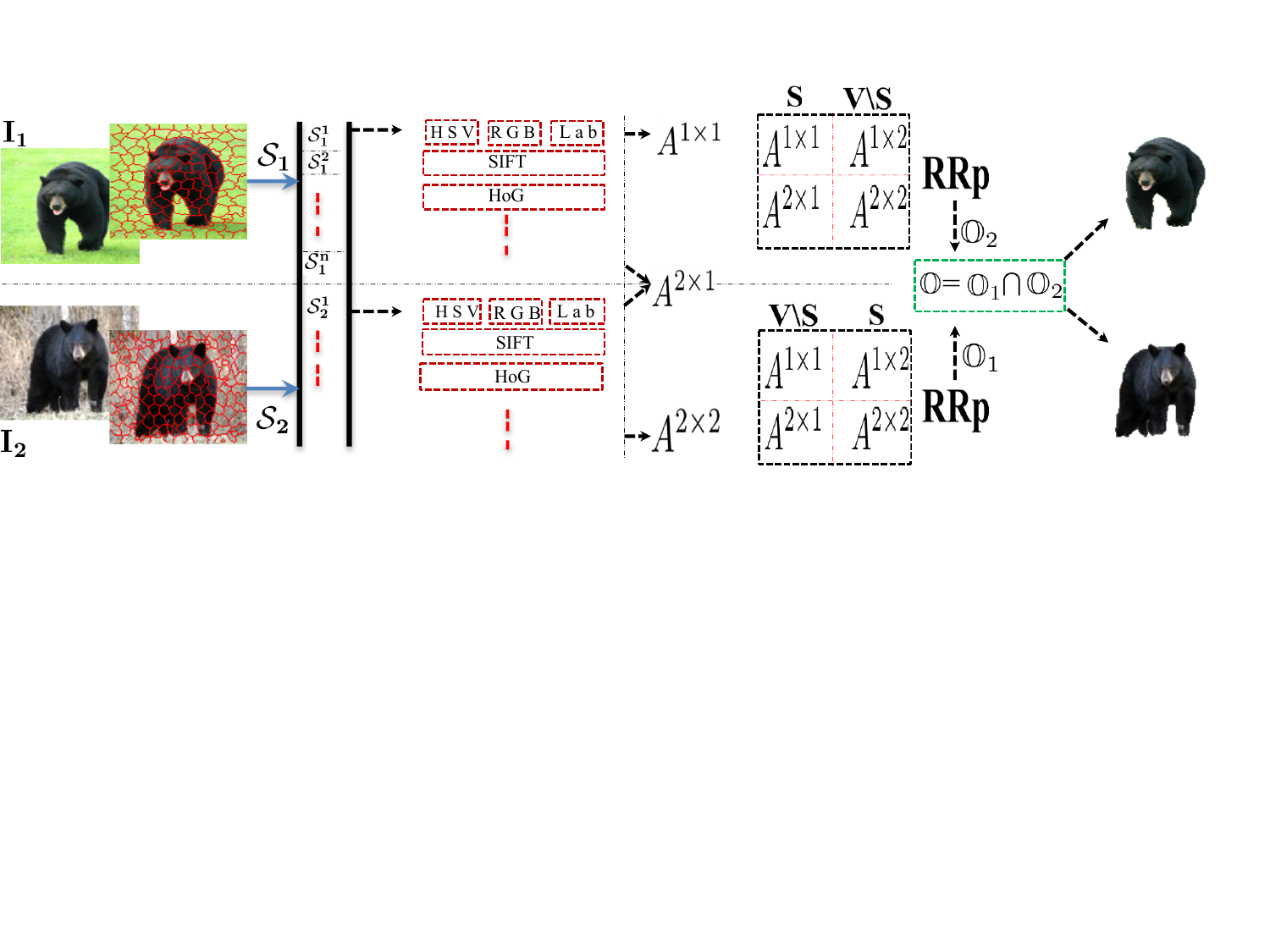}
	\caption{\small Overview of our unsupervised co-segmentation algorithm.}
	\label{fig:FrameworkCo-Seg}
\end{figure*}

\begin{figure*}[t]
	\centering
	\includegraphics[width=1\linewidth ,trim=0cm 3cm 0cm 1cm,clip]{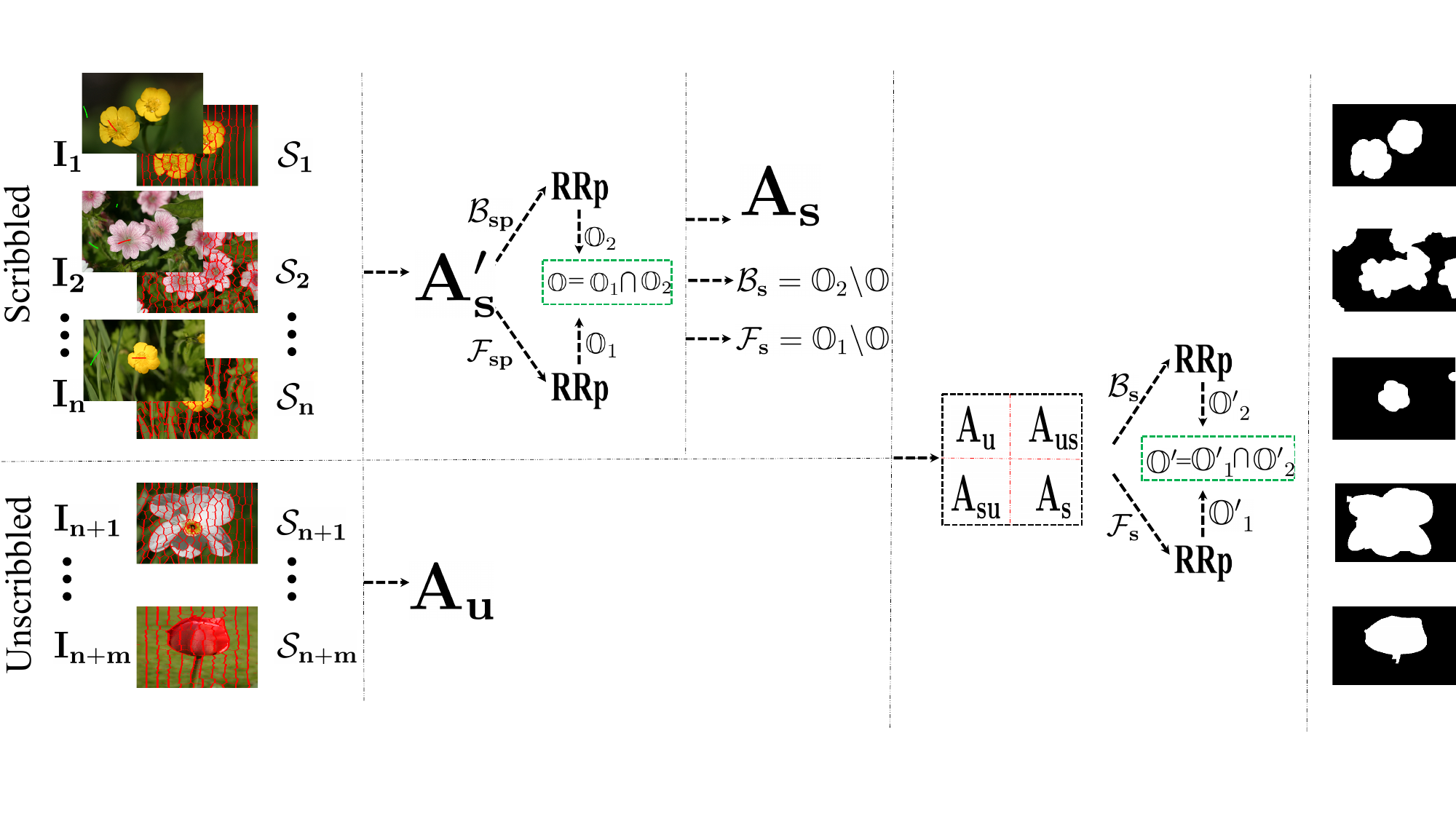}
	\caption{\small Overview of our interactive co-segmentation algorithm.} 
	\label{fig:interactiveco-segmentationworkflow}
\end{figure*}

Figure \ref{fig:FrameworkCo-Seg} and  \ref{fig:interactiveco-segmentationworkflow} show the pipeline of our unsupervised
and interactive co-segmentation algorithms, respectively. 

In figure \ref{fig:FrameworkCo-Seg}, $\vct{I}_{1}$ and $\vct{I}_{2}$ are the given pair of images while $\mathcal{S}_{1}$ and $\mathcal{S}_{2}$ represent the corresponding sets of superpixels. The affinity is built using the objectness score of the superpixels and using different handcrafted features extracted from the superpixels. The set of nodes $V$ is then divided into two as the constraint set ($S$) and the non-constraint ones, $V\backslash S$. We run the CDS algorithm twice: first, setting the nodes of the graph that represent the first image as constraint set and $\mathbb{O}_2$ represents our output. Second we change the constraint set $S$ with nodes that come from the second image and $\mathbb{O}_1$ represents the output. The intersection $\mathbb{O}$ refines the two results and represents the final output of the proposed unsupervised co-segmentation approach.

Our interactive co-segmentation approach, as shown using Figure \ref{fig:interactiveco-segmentationworkflow}, needs user interaction which guides the segmentation process putting scribbles (only) on some of the images with ambiguous objects or background. $\vct{I}_{1},\vct{I}_{2},...\vct{I}_{n}$ are the scribbled images and $\vct{I}_{n+1}, ..., \vct{I}_{n+m}$ are unscribbled ones. The corresponding sets of superpixels are represented as $\mathcal{S}_{1},\mathcal{S}_{2},...\mathcal{S}_{n},...\mathcal{S}_{n+1},...\mathcal{S}_{n+m}$. $\vct{A'}_\vct{s}$ and $\vct{A}_\vct{u}$ are the affinity matrices built using handcrafted feature-based similarities among superpixels of scribbled and unscribbled images respectively. Moreover, the affinities incorporate the objectness score of each node of the graph. $\mathcal{B}_{\vct{s}\vct{p}}$ and $\mathcal{F}_{\vct{s}\vct{p}}$ are (respectively) the background and foreground superpixels based on the user provided information. The CDS algorithm is run twice over $\vct{A'}_\vct{s}$ using the two different user provided information as constraint sets which results outputs $\mathbb{O}_1$ and $\mathbb{O}_2$. The intersection of the two outputs, $\mathbb{O}$, help us get new foreground and background sets represented by $\mathcal{B}_\vct{s}$, $\mathcal{F}_\vct{s}$. Modifying the affinity $\vct{A'}_\vct{s}$, putting the similarities among elements of the two sets to zero, we get the new affinity $\vct{A}_\vct{s}$. We then build the biggest affinity which incorporates all images' superpixels. As our affinity is symmetric, $\vct{A}_{\vct{u}\vct{s}}$ and $\vct{A}_{\vct{s}\vct{u}}$ are equal and incorporates the similarities among the superpixels of the scribbled and unscribbled sets of images. Using the new background and foreground sets as two different constraint sets, we run CDS twice which results outputs $\mathbb{O'}_1$ and $\mathbb{O'}_2$ whose intersection ($\mathbb{O'}$) represents the final output.

\subsection{Experiments and results}

Given an image, we over-segment it to get its superpixels  $\mathcal{S}$, which are considered as vertices of a graph. We then extract different features from each of the superpixels. The first features which we consider are features from the different color spaces: RGB, HSV and CIE Lab. Given the superpixels, say size of $n$, of an image $i$, $\mathcal{S}_i$, $\mathcal{F}_c^i$ is a matrix of size $n \times 9$ which is the mean of each of the channels of the three color spaces of pixels of the superpixel. The mean of the SIFT features extracted from the superpixel $\mathcal{F}_s^i$ is our second feature. The last feature which we have considered is the rotation invariant histogram of oriented gradient (HoG), $\mathcal{F}_h^i$. 

The dot product of the SIFT features  is considered as the SIFT similarity among the nodes, let us say the corresponding affinity matrix is $A_s$. Motivated by \cite{MorML2016}, the similarity among the nodes of image $i$ and image $j$ ($i\neq j$), based on color, is computed from their Euclidean distance $\mathcal{D}_c^{i\times j}$ as 

\[ A_c^{i\times j} = max(\mathcal{D}_c) - \mathcal{D}_c^{i\times j} + min(\mathcal{D}_c)\]

The HoG similarity among the nodes, $A_h^{i\times j}$, is computed in a similar way , as $A_c$, from the diffusion distance. All the similarities are then min max normalized.

We then construct the  $A_c^{i\times i}$, the similarities among superpixels of image $i$,  which only considers adjacent superpixels as follows. First, construct the dissimilarity graph using their Euclidean distance considering their average colors as weight. Then, compute the geodesic distance as the accumulated edge weights along their shortest path on the graph, we refer the reader to \cite{ZemeneP15} to see how such type of distances improve the performance of dominant sets. Assuming the computed geodesic distance matrix is $\mathcal{D}_{geo}$, the weighted edge similarity of superpixel $p$ and superpixel $q$, say $e_{p,q}$, is computed as

\begin{equation}
e_{p,q} = 
\begin{cases}
0, \hspace*{1cm}\text{if\quad p and q are not adjacent}, & \\
max(\mathcal{D}_{geo}) - \mathcal{D}_{geo}(p,q)+ min(\mathcal{D}_{geo}),&\text{otherwise}
\end{cases}
\end{equation}

$A_h^{i\times i}$ for HoG is computed in a similar way while and $A_s^{i\times i}$ for SIFT is built by just keeping adjacent edge similarities.

Assuming we have $I$ images, the final affinity $A_\gamma$ ($\gamma$ can be $c$, $s$ or $h$ in the case of color, SIFT or HOG respectively) is built as

$$
A_\gamma = 
\begin{pmatrix} 
A_\gamma^{1 \times 1} & .  .  & A_\gamma^{1 \times j} & . . & A_\gamma^{1 \times I} \\ 
. & ~~ . ~~ &  .  & . & .\\
~~ ~~ & ~~ ~~ & ~~ ~~ &   \\
A_\gamma^{j \times 1} & .  .  & A_\gamma^{j \times j} & . & A_\gamma^{1 \times I}\\ 
. & ~~ ~~ &  . & . & .\\
~~ ~~ & ~~ ~~ & ~~ ~~\\
A_\gamma^{I \times 1} & .  .  & A_\gamma^{I \times j} & . . & A_\gamma^{I \times I}\\ 
\end{pmatrix}
$$ 

As our goal is to segment common foreground objects out, we should consider how related backgrounds are eliminated. As shown in the examplar image pair of Figure \ref{fig:ImgPairExemplar} (top right), the two images have a related background to deal with it which otherwise would be included as part of the co-segmented objects.
To solve this problem we borrowed the idea from  \cite{WanShuYicJiaCVPR2014} which proposes a robust background
measure, called boundary connectivity. Given a superpixel $\mathcal{SP}_i$, it computes, based on the background measure, the backgroundness probability $\mathcal{P}_b^i$. We compute the probability of the superpixel being part of an object $\mathcal{P}_f^i$ as its additive inverse, $\mathcal{P}_f^i$ = 1 - $\mathcal{P}_b^i$. From the probability $\mathcal{P}_f$ we built a score affinity $A_m$ as 

\[A_m(i,j) = \mathcal{P}_f^i* \mathcal{P}_f^j\]

\subsubsection{Optimization}

We model the foreground object extraction problem as the optimization of the similarity values among all image superpixels.
The objective utility function is designed to assign the object region a membership score of greater than zero and the background region zero membership score, respectively. The optimal object region is then obtained by maximizing the utility function. Let the membership score of $N$ superpixels be $\{x_i\}_{i=1}^N$, the $(i,j)$ entry of a matrix $A_z$ is $z_{ij}$. Our utility function, combining all the aforementioned terms ($A_c$,$A_s$,$A_h$ and $A_m$), is thus defined, based on equation (\ref{eqn:parQP}), as:

\begin{equation}
\sum\limits_{i=1}^{N} \sum\limits_{j=1}^{N} \left\{ \frac{1}{2}\underbrace{ x_ix_jm_{ij}}_\text{objectness score} +    \frac{1}{6}x_ix_j\underbrace{\left(c_{ij} + s_{ij} + h_{ij} \right)}_\text{feature similarity} - \alpha x_ix_j \right\}
\end{equation}

The parameter $\alpha $ is fixed based on the (non-)constraint set of the nodes. For the case of unsupervised co-segmentation, the nodes of the pairs of images are set (interchangeably) as constraint set where the intersection of the corresponding results give us the final co-segmented objects.

In the interactive setting, every node $i$ (based on the information provided by the user) has three states: $i \in FGL $, ($i$ is labeled as foreground label), $i \in BGL$ ( $i$ is labeled as background label) or $i \in V\backslash ({FGL \cup BGL})$ ($i$ is unlabeled).
Hence, the affinity matrix $A=(a_{ij})$ is modified by setting $a_{ij}$ to zero if nodes $i$ and $j$ have different labels
(otherwise we keep the original value).

\begin{figure}
	\centering
	\includegraphics[width=0.9\linewidth,trim=3cm 2cm 0cm 0cm,clip]{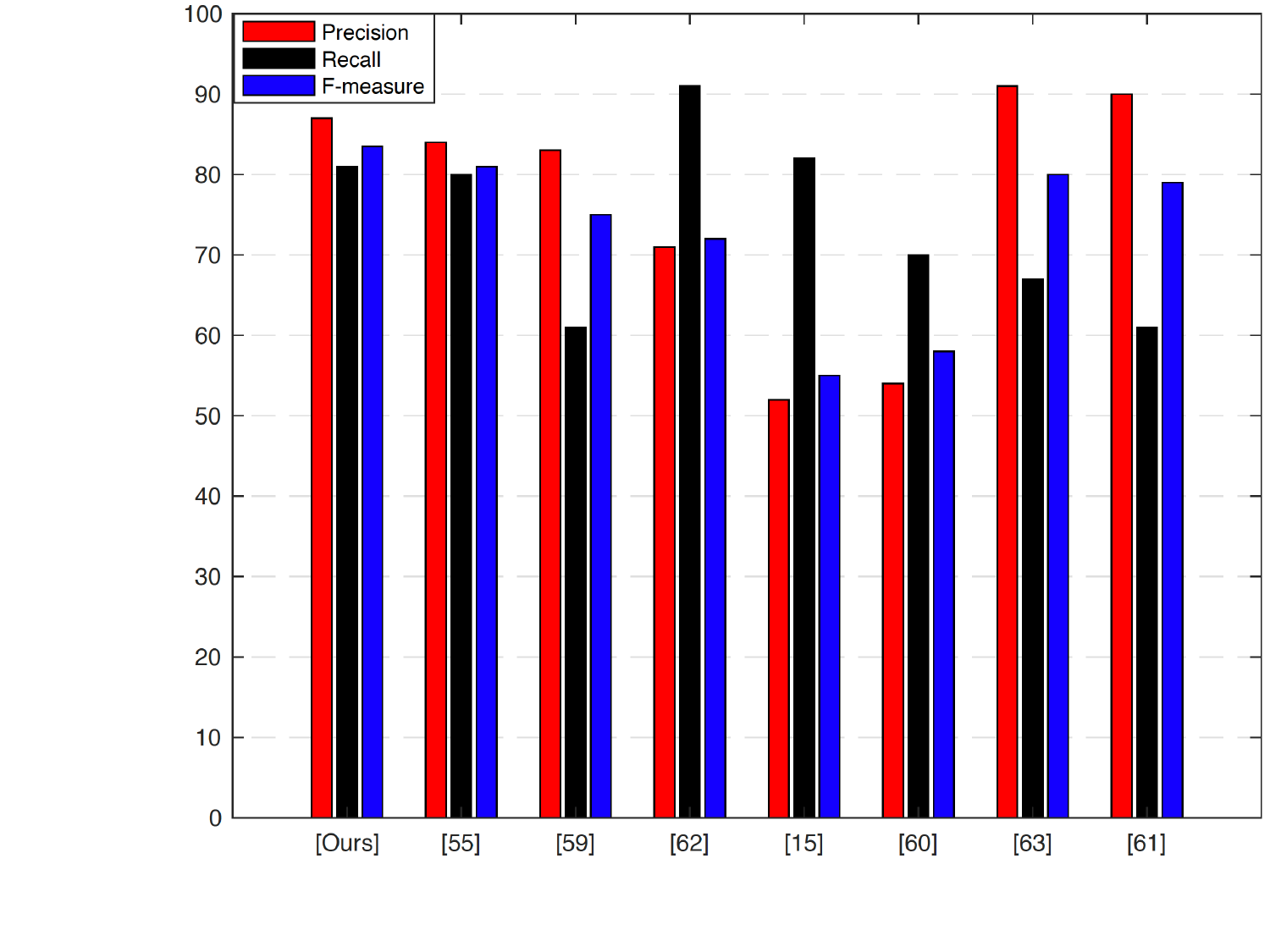}
	\caption{\small Precision, Recall and F-Measure based performance comparison of our unsupervised co-segmentation method with the state-of-the art approaches on image pair dataset}
	\label{fig:Imagepairesult}
\end{figure}

The optimization, for both cases, is represented in the pipelines by '\textbf{RRp}' (replicator dynamics).

 \begin{figure*}
 	\centering
 	\includegraphics[width=0.9\linewidth]{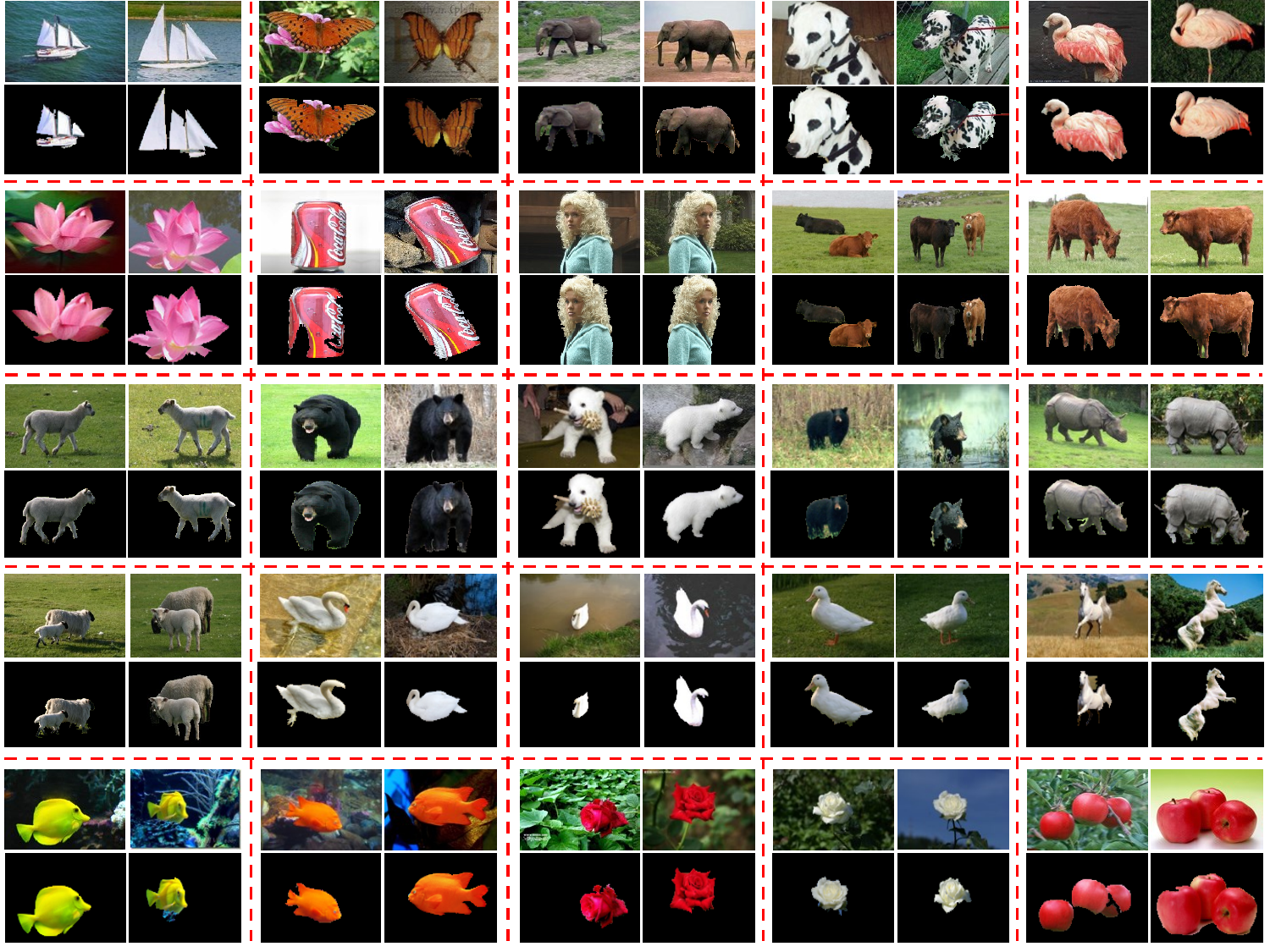}
 	\caption{\small Examplar qualitative results of our unsupervised method tested on image pair dataset. \textbf{Upper row:} Original image \textbf{Lower row:} Result of the proposed unsupervised algorithm.}
 	\label{fig:ExamplarResultsCoSeg}
 \end{figure*}

To evaluate the performance of our algorithms, we conducted extensive experiments on standard benchmark datasets that are widely used to evaluate the co-segmentation problem: image pairs \cite{HonKinPAMI2011} and MSRC \cite{MicArmJohCeCVPR2013}. The image pairs dataset consists 210 images (105 image pairs) of different animals, flowers, human objects, buses, etc. Each of the image pairs contains one or more similar objects. Some of them are relatively simple and some other contains set of complex image pairs, which contain foreground objects with higher appearance variations or low contrast objects with complex backgrounds. 

MSRC dataset has been widely used to evaluate the performance of image co-segmentation methods. It contains 14 categories with 418 images in total. We evaluated our interactive co-segmentation algorithm on nine selected object classes of MSRC dataset (bird, car, cat, chair, cow, dog, flower, house, sheep), which  contains 25\texttildelow30 images per class. We put foreground and background scribbles on 15\texttildelow20 images per class. Each image was over-segmented to 78\texttildelow83 SLIC superpixels using the VLFeat toolbox.

As customary, we measured the performance of our algorithm using precision, recall and F-measure, which were computed based on the output mask and human-given segmentation ground-truth. Precision is calculated as the ratio of correctly detected objects to the number of detected object pixels, while recall is the ratio of correctly detected object pixels to the number of ground truth pixels. We have computed the F-measure by setting $\gamma^2$ to 0.3 as used in \cite{HonKinPAMI2011}\cite{HuaXiaZhuTIP2013}\cite{AvikChaVelECCV16}.

\begin{figure}
	\centering
	\includegraphics[width=0.9\linewidth,trim=3.5cm 1.5cm 0cm 0cm,clip]{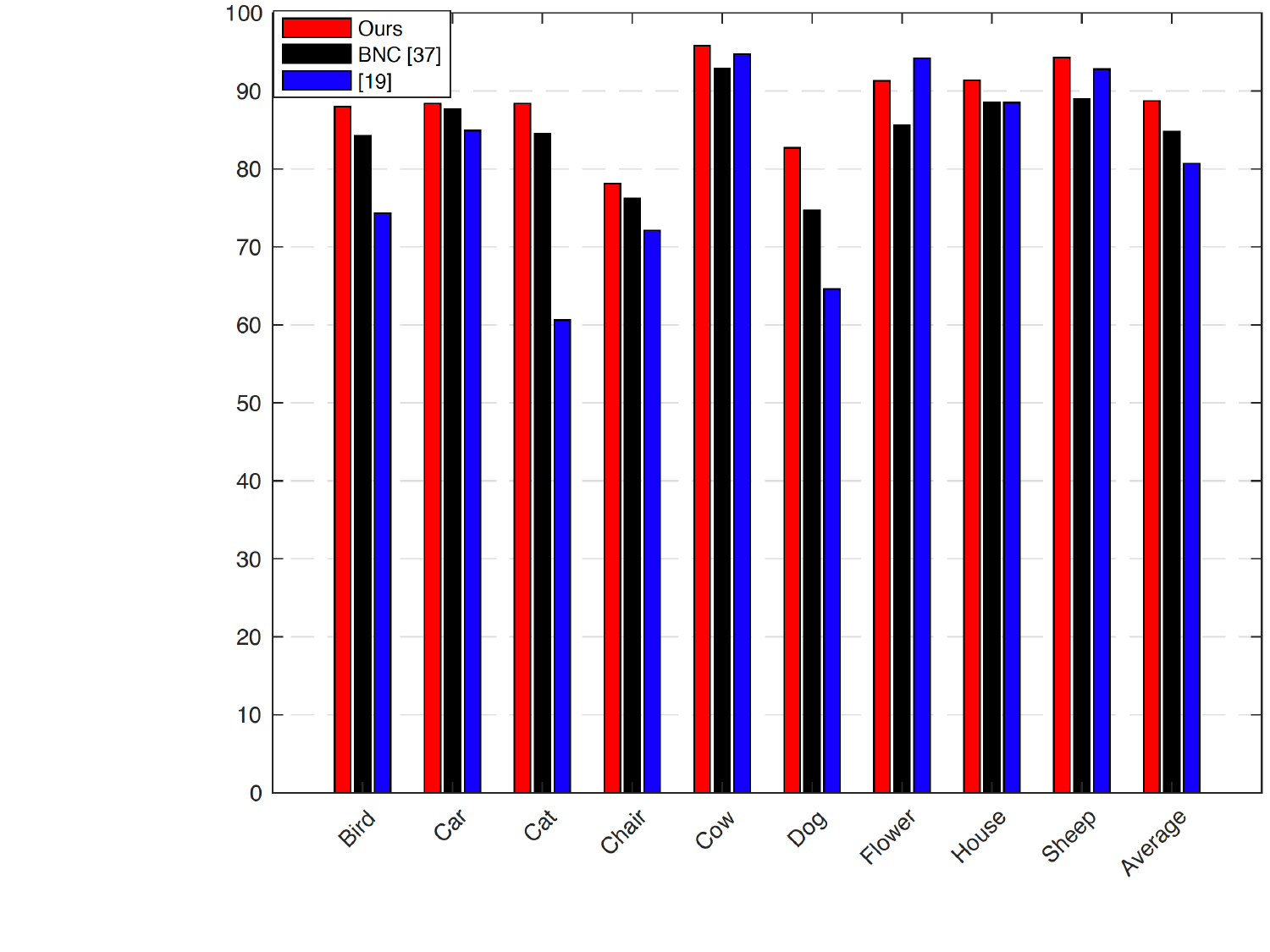}
	\caption{\small F-Measure based performance Comparison of our interactive co-segmentation method with state-of-the-art methods on MSRC dataset.}
	\label{fig:MSRC_Result}
\end{figure}

We have applied Biased Normalized Cut (BNC) \cite{MajiNishVishMalCVPR11}
on co-segmentation problem on MSRC dataset by using the same similarity matrix we used to test our method, and the comparison result of each object class is shown in Figure \ref{fig:MSRC_Result}. As can be seen, our method significantly surpasses BNC and  \cite{XinJiaLinMinTIP2015} in average F-measure. Furthermore, we have tested our interactive co-segmentation method, BNC and  \cite{XinJiaLinMinTIP2015} on image pairs dataset by putting scribbles on one of the two images. As can be observed from Table \ref{table:ImagePair}, our algorithm substantially outperforms BNC and \cite{XinJiaLinMinTIP2015} in precision and F-measure
(the recall score being comparable among the three competing algorithms). 

In addition to that, we have examined our unsupervised co-segmentation algorithm by using image pairs dataset, the barplot in Figure \ref{fig:Imagepairesult} shows the quantitative result of our algorithm comparing to the state-of-the-art methods \cite{AvikChaVelECCV16}\cite{ChWoJaChaCVPR2015}\cite{XiaZhiqBaTIP2014}. As shown here, our algorithm achieves the best F-measure comparing to all other state-of-the-art methods. The qualitative performance of our unsupervised algorithm is shown in Figure \ref{fig:ExamplarResultsCoSeg} on some example images taken from image pairs dataset. As can be seen, Our approach can effectively detect and segment the common object of the given pair of images.

\begin{table}[bth]
	
	\begin{center}
		\begin{tabular}{lllllllllllllll} 
			\hline\noalign{\smallskip}
			Metrics & $ Precision$  & $ Recall$ & $F-measure$ \\
			\noalign{\smallskip}
			\hline
			\noalign{\smallskip}
			\cite{XinJiaLinMinTIP2015}   & 0.5818 &  0.8239 &  0.5971 \\    
			BNC & 0.6421 & \textbf{0.8512} &  0.6564\\ 
			Ours & \textbf{0.7076} & 0.8208 &  \textbf{0.7140}\\ 	\hline	
		\end{tabular}
	\end{center}
	\caption{\small Results of our interactive co-segmentation method on Image pair dataset putting user scribble on one of the image pairs}
	\label{table:ImagePair}
\end{table}

\section{Conclusions}

In this paper, we have introduced the notion of a {\em constrained dominant set} and have demonstrated its applicability
to problems such as interactive image segmentation and co-segmentation (in both the unsupervised and the interactive flavor).
In our perspective, these can be thought of as ``constrained'' segmentation problems involving an external source of information (being it, for example, a user annotation or a collection of related images to segment jointly) which somehow drives the whole segmentation process. The approach is based on some properties of a family of quadratic optimization problems related to dominant sets which show that, by properly selecting a regularization parameter that controls the structure of the underlying function, we are able to ``force''€ all solutions to contain the constraint elements. 
The proposed method is flexible and is capable of dealing with various forms of constraints and input modalities, such as 
scribbles and bounding boxes, in the case of interactive segmentation. 
Extensive experiments on benchmark datasets have shown that our approach considerably improves the state-of-the-art results on the problems addressed. This provides evidence that constrained dominant sets hold promise as a powerful and principled framework to address a large class of computer vision problems formulable in terms of constrained grouping. Indeed, we mention that they are already being used successfully in other applications such as content-based image retrieval \cite{ZemAlePel16}, multi-target tracking \cite{TesfayeZPPS17} and image geo-localization \cite{EyaYonHarAndMarMubPAMI}.

\bigskip
\noindent
{\bf Acknowldegments}. This work has been partly supported by the Samsung Global Research Outreach Program.

\bibliographystyle{IEEETran} 

\bibliography{EyasuPami-2.bib}

\newpage
\vspace*{-10cm}
\begin{IEEEbiography}[{\includegraphics[width=1.25in,height=1.25in,,keepaspectratio]{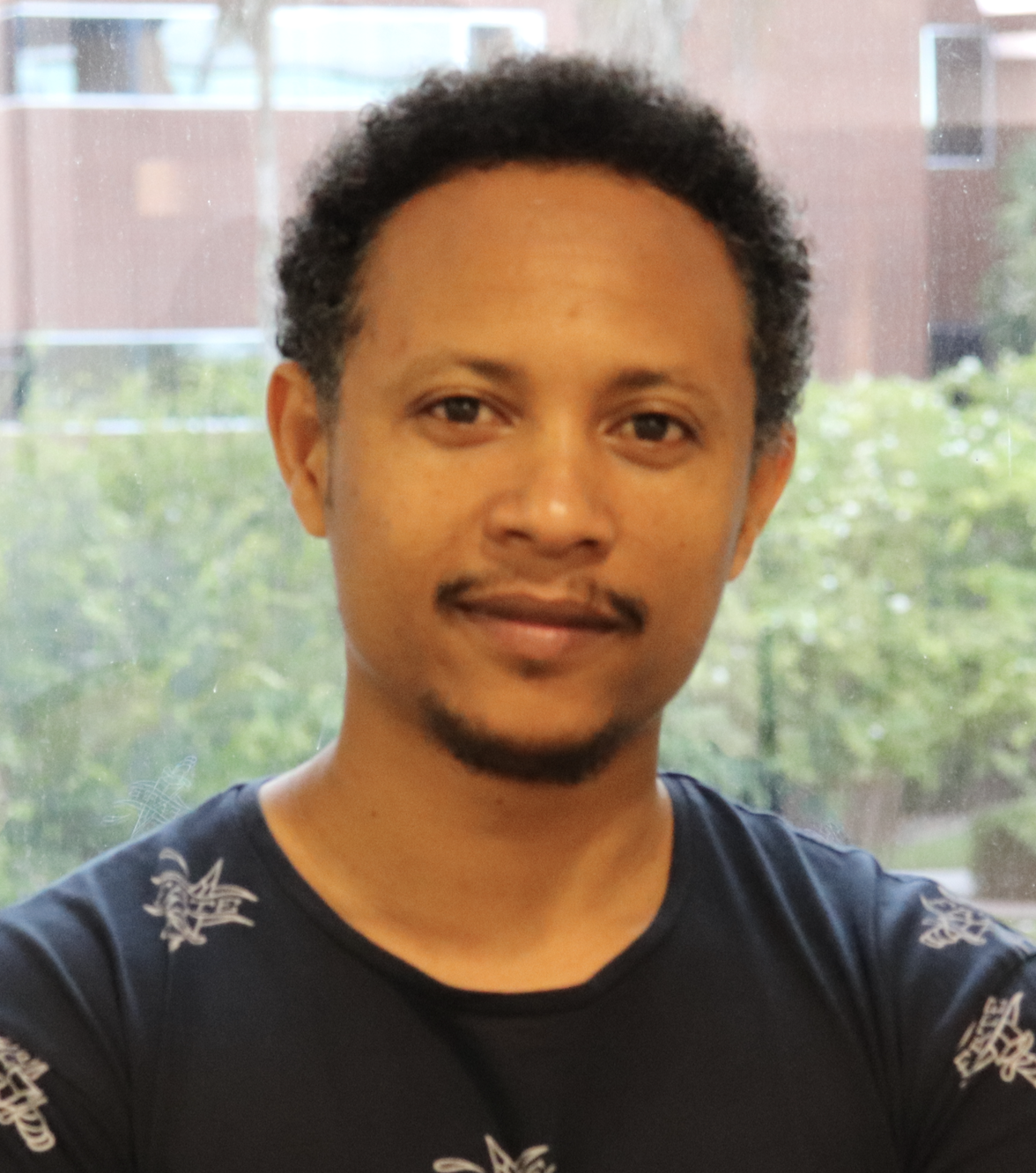}}]{Eyasu Zemene}
	received the BSc degree in Electrical Engineering from Jimma University in 2007, he then worked at Ethio Telecom for 4 years till he joined Ca' Foscari University (October 2011) where he got his MSc in Computer Science in June 2013. September 2013, he won a 1 year research fellow to work on Adversarial Learning at Pattern Recognition and Application lab of University of Cagliari. Since September 2014 he is a PhD student of CaFoscari University under the supervision of prof. Pelillo. Working towards his Ph.D. he is trying to solve different computer vision and pattern recognition problems using theories and mathematical tools inherited from graph theory, optimization theory and game theory. Currently, Eyasu, as part of his PhD, is working as a research assistant at Center for Research in Computer Vision at University of Central Florida under the supervision of Dr. Mubarak Shah. His research interests are in the areas of Computer Vision, Pattern Recognition, Machine Learning, Graph theory and Game theory.
\end{IEEEbiography}\vspace*{-8cm}

\begin{IEEEbiography}[{\includegraphics[width=1.1in,height=1.5in,clip,keepaspectratio]{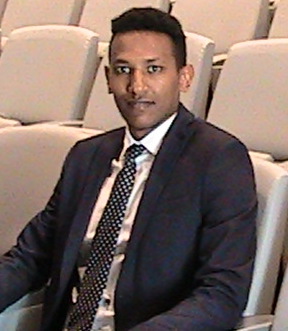}}]{Leulseged Tesfaye}Received his BSc in computer science from Jimma University in 2012.  After working for two years as EUC engineer at Kifiya financial Technology he joined Ca' Foscari University of Venice where he received his MSc in computer science in June 2016. He is Currently working towards his PhD degree at Ca' Foscari University of Venice, Italy under the supervision of prof. Marcello Pelillo . His research interest includes Computer Vision, Pattern Recognition, Machine Learning, Game Theory and Graph Theory.
\end{IEEEbiography} \vspace*{-8cm}

\begin{IEEEbiography}[{\includegraphics[width=1.1in,height=1.5in,clip,keepaspectratio]{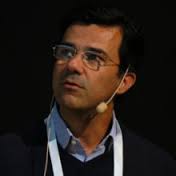}}]{Marcello Pelillo} is Professor of Computer Science at Ca' Foscari University in Venice, Italy, where he directs the European Centre for Living Technology (ECLT) and the Computer Vision and Pattern Recognition group. He held visiting research positions at Yale University, McGill University, the University of Vienna, York University (UK), the University College London, the National ICT Australia (NICTA), and is an Affiliated Faculty Member of Drexel University, Department of Computer Science. He has published more than 200 technical papers in refereed journals, handbooks, and conference proceedings in the areas of pattern recognition, computer vision and machine learning. He is General Chair for ICCV 2017, Track Chair for ICPR 2018, and has served as Program Chair for several conferences and workshops, many of which he initiated (e.g., EMMCVPR, SIMBAD, IWCV). He serves (has served) on the Editorial Boards of the journals IEEE Transactions on Pattern Analysis and Machine Intelligence (PAMI), Pattern Recognition, IET Computer Vision, Frontiers in Computer Image Analysis, Brain Informatics, and serves on the Advisory Board of the International Journal of Machine Learning and Cybernetics. Prof. Pelillo has been elected a Fellow of the IEEE and a Fellow of the IAPR, and has recently been appointed IEEE SMC Distinguished Lecturer. His Erd\"os number is 2.
\end{IEEEbiography}\vspace*{-5cm}

\end{document}